\documentclass[11pt,a4paper]{article} 

\usepackage[margin=1in]{geometry}

\usepackage{microtype}
 \usepackage{booktabs}  

\usepackage[utf8]{inputenc} 
\usepackage[T1]{fontenc}    
\usepackage{lmodern}
\usepackage{url}            
\usepackage{booktabs}       
\usepackage{amsfonts}       
\usepackage{nicefrac}       
\usepackage{microtype}      
\usepackage{amssymb,amsthm,amsmath}
\usepackage{color}
\usepackage[usenames,dvipsnames,svgnames,table]{xcolor}
\usepackage[colorlinks=true, linkcolor=red, urlcolor= MidnightBlue, citecolor=gray]{hyperref}
\usepackage{graphicx}
\graphicspath{{./figures/}}
\usepackage{epstopdf}
\usepackage{float}
\usepackage{caption}
\usepackage{bbm}
\usepackage{subfig}
\usepackage{bm}
\usepackage{algorithm}
\usepackage{algpseudocode}
\usepackage{dblfloatfix} 
\usepackage{natbib}

  \usepackage{nth}
  \usepackage{intcalc}

\newtheorem{theorem}{Theorem}

\newtheorem{definition}{Definition}
\newtheorem{problem}[definition]{Problem}

\newcommand{\tr}{\mathrm{tr}}

\newcommand{\norm}[1]{\|#1\|}

\newcommand{\hide}[1]{}
\newcommand{\spara}[1]{
\medskip \noindent {\bf #1}}  

\title{DeepWalking Backwards:\\ From Embeddings Back to Graphs}

\author{
Sudhanshu Chanpuriya \thanks{UMass Amherst. \texttt{schanpuriya@umass.edu}} \and 
Cameron Musco \thanks{UMass Amherst. \texttt{cmusco@cs.umass.edu}} \and 
Konstantinos Sotiropoulos \thanks{Boston University. \texttt{ksotirop@bu.edu}} \and 
Charalampos E. Tsourakakis\thanks{ISI Foundation \& Boston University. \texttt{tsourolampis@gmail.com }}
}
\date{}

\begin{document}
   \maketitle 

\begin{abstract}

Low-dimensional node embeddings play a key role in analyzing graph datasets.
However, little work studies exactly \emph{what information is encoded} by popular embedding methods, and how this information correlates with performance in downstream learning tasks.
We tackle this question by studying whether embeddings can be \emph{inverted} to (approximately) recover the graph used to generate them.
Focusing on a variant of the popular DeepWalk method \citep{PerozziAl-RfouSkiena:2014, QiuDongMa:2018}, we present algorithms for accurate embedding inversion -- i.e., from the low-dimensional embedding of a graph $G$, we can find a graph $\tilde G$ with a very similar embedding. We perform numerous experiments on real-world networks, observing that significant information about $G$, such as specific edges and bulk properties like triangle density, is often lost in $\tilde G$. However, community structure is often preserved or even enhanced. Our findings are a step towards a more rigorous understanding of exactly what information embeddings encode about the input graph, and why this information is useful for learning tasks.

\end{abstract}

\section{Introduction}
\label{sec:intro}


Low-dimensional node embeddings are a primary tool in graph mining and machine learning. They are used for node classification, community detection, link prediction, and graph generative models. 
Classic approaches like spectral clustering \citep{shi2000normalized,ng2002spectral}, Laplacian eigenmaps \citep{belkin2003laplacian}, IsoMap \citep{tenenbaum2000global}, and  locally linear embeddings \citep{roweis2000nonlinear} use spectral embeddings derived for the graph Laplacian, adjacency matrix, or their variants. Recently, neural-network and random-walk-based embeddings have become popular due to their superior performance in many settings. Examples include DeepWalk \citep{PerozziAl-RfouSkiena:2014}, node2vec \citep{grover2016node2vec}, LINE \citep{tang2015line}, NetMF \citep{QiuDongMa:2018}, and  many others \citep{cao2016deep,kipf2016semi,wang2016structural}.  In many cases, these methods can be viewed as variants on classic spectral methods, producing an approximate low-dimensional factorization of an implicit matrix representing graph structure \citep{QiuDongMa:2018}.

\spara{Problem definition.} We focus on the following high-level question: 

\begin{quotation}
\noindent What graph properties are encoded in and can be recovered from node embeddings? How do these properties correlate with learning tasks? 
\end{quotation}

\noindent We study the above question on undirected graphs with non-negative edge weights. Let $\mathcal{G}$ denote the set of all such graphs with $n$ nodes. We formalize the question via Problems~\ref{prob:inversion} and \ref{prob:stab} below. 

\begin{problem}[Embedding Inversion]\label{prob:inversion}  Given an embedding algorithm $\mathcal{E}: \mathcal{G} \rightarrow \mathbb{R}^{n \times k}$ and the embedding $\mathcal{E}(G)$ for some $G \in \mathcal{G}$, produce $\tilde G \in \mathcal{G}$ with $\mathcal{E}(\tilde G) = \mathcal{E}(G)$ or such that $\norm{\mathcal{E}(\tilde G) - \mathcal{E}(G)}$ is small for some norm $\norm{\cdot}$.
\end{problem}

\noindent We refer to $k$ as the {\em embedding dimension}. A solution to Problem \ref{prob:inversion}  lets us approximately invert the embedding $\mathcal{E}(G)$ to obtain a graph. It is natural to ask what structure is common between  $G, \tilde G$. Using the same notation as Problem~\ref{prob:inversion}, our second problem is as follows. 

\begin{problem}[Graph Recovery]\label{prob:stab}   
Given $G, \tilde G$ such that $\norm{\mathcal{E}(\tilde G) - \mathcal{E}(G)}$  is small for some matrix norm $\norm{\cdot}$, how close are $G,\tilde G$  in terms of common edges, degree sequence, triangle counts, and community structure?
\end{problem}

Answering Problems~\ref{prob:inversion} and ~\ref{prob:stab} is an important step towards a better understanding of a node embedding method $\mathcal{E}$.  We focus on the popular DeepWalk method of \citet{PerozziAl-RfouSkiena:2014}. DeepWalk embedding 
can be interpreted as low-rank approximation of a pointwise mutual information (PMI) matrix based on node co-occurrences in random walks \citep{GoldbergLevy:2014}. The NetMF method of \citet{QiuDongMa:2018} directly implements this low-rank approximation using  SVD, giving a variant with improved performance in many tasks. Due to its mathematically clean definition, we use this variant. Many embedding methods can be viewed similarly -- as producing a low-rank approximation of some graph-based similarity matrix.  We expect our methods to extend to such embeddings.

\begin{figure*}
\centering
\begin{tabular}{cc}
\includegraphics[width=0.40\linewidth]{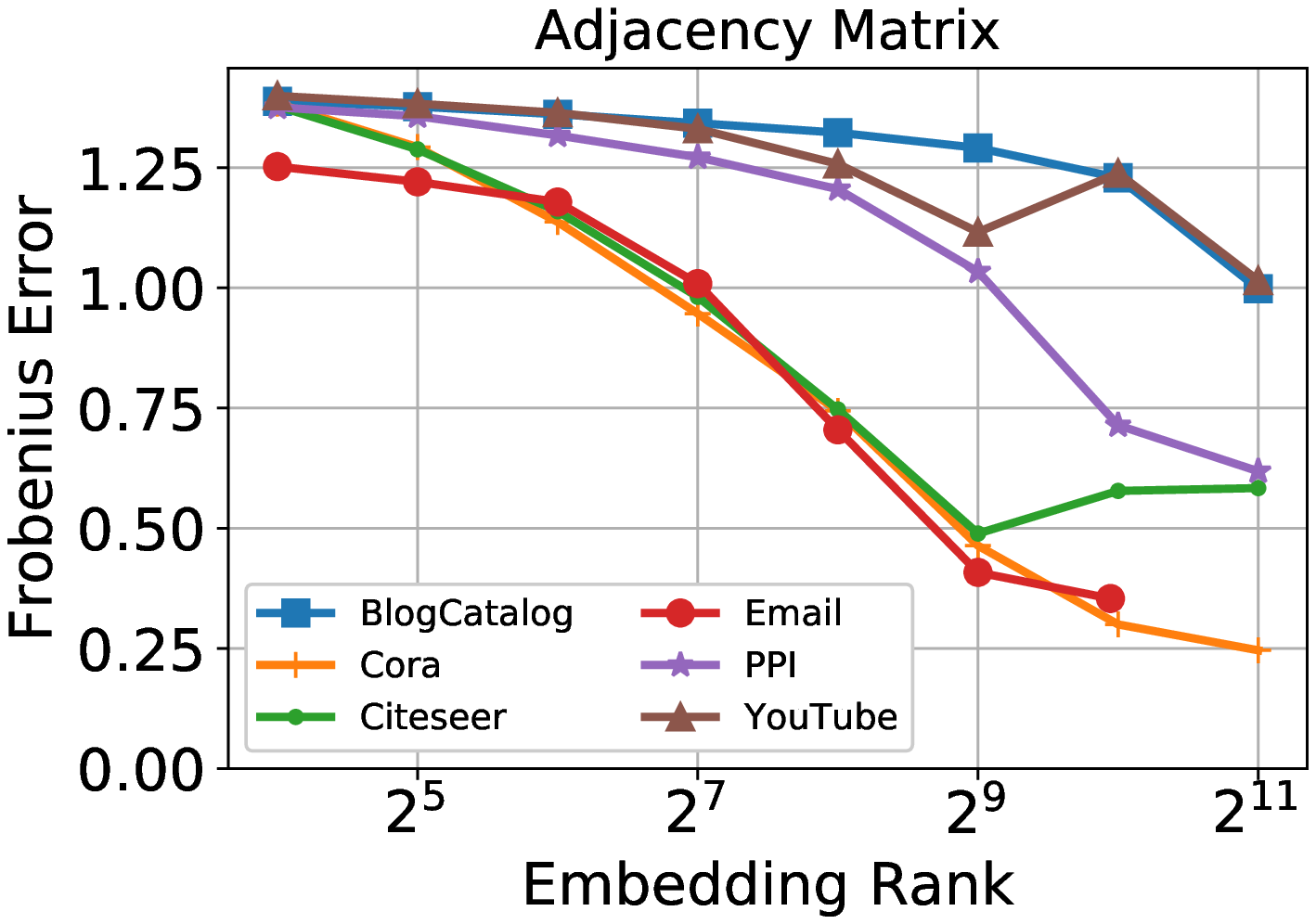} & 
\includegraphics[width=0.40\linewidth]{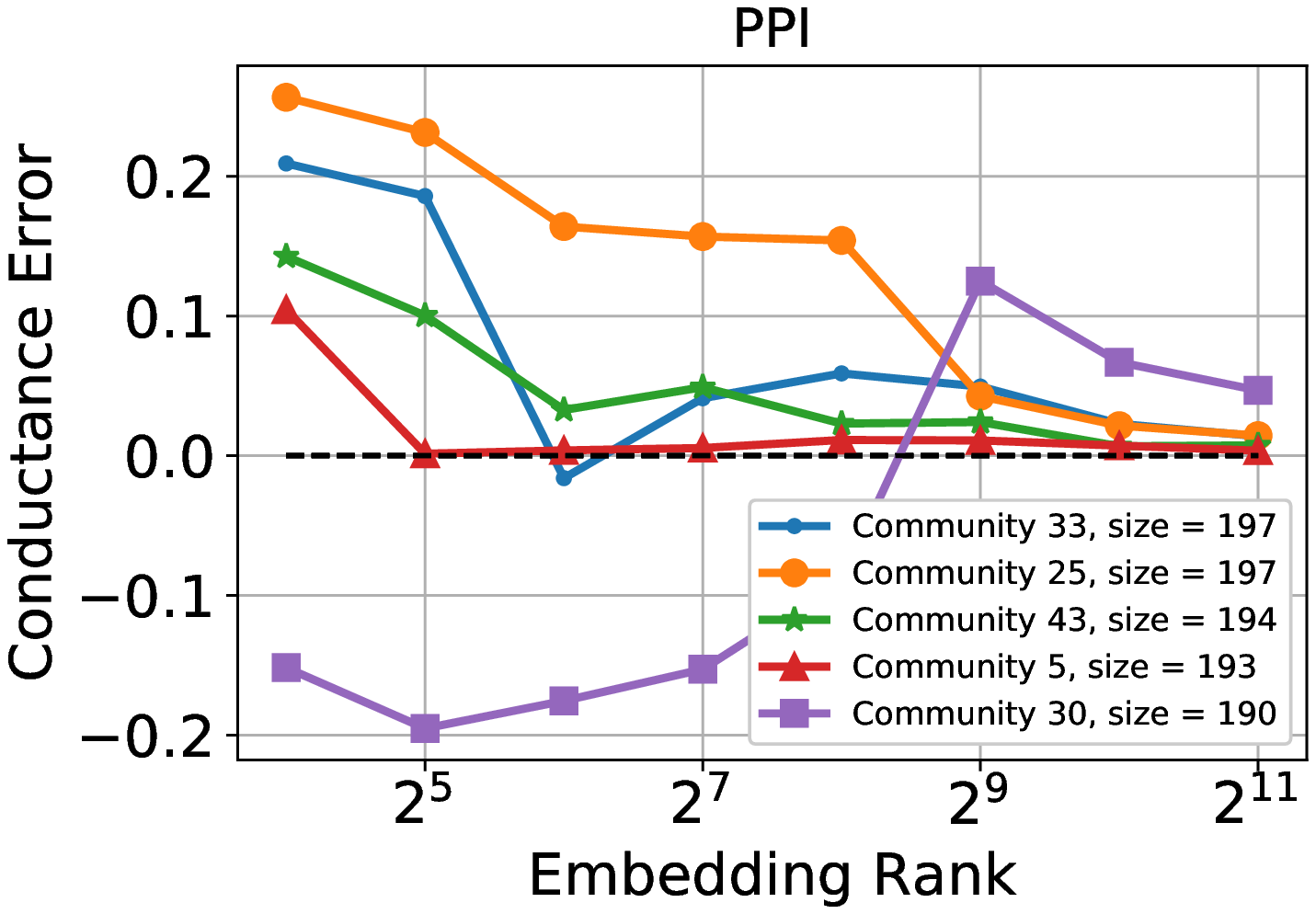} \\
(a) & (b) 
\end{tabular}
\caption{\label{fig:intro_fig} (a) Relative Frobenius error $\norm{A-\tilde A}_F/\norm{A}_F$ between the adjacency matrices of $G$ and $\tilde G$. (b) Relative error between $G$ and $\tilde{G}$ for the conductances of the five largest communities (corresponding to biological states) in a human protein-protein interaction network.}
\end{figure*}

\begin{figure}
\centering
\includegraphics[width=0.33\linewidth]{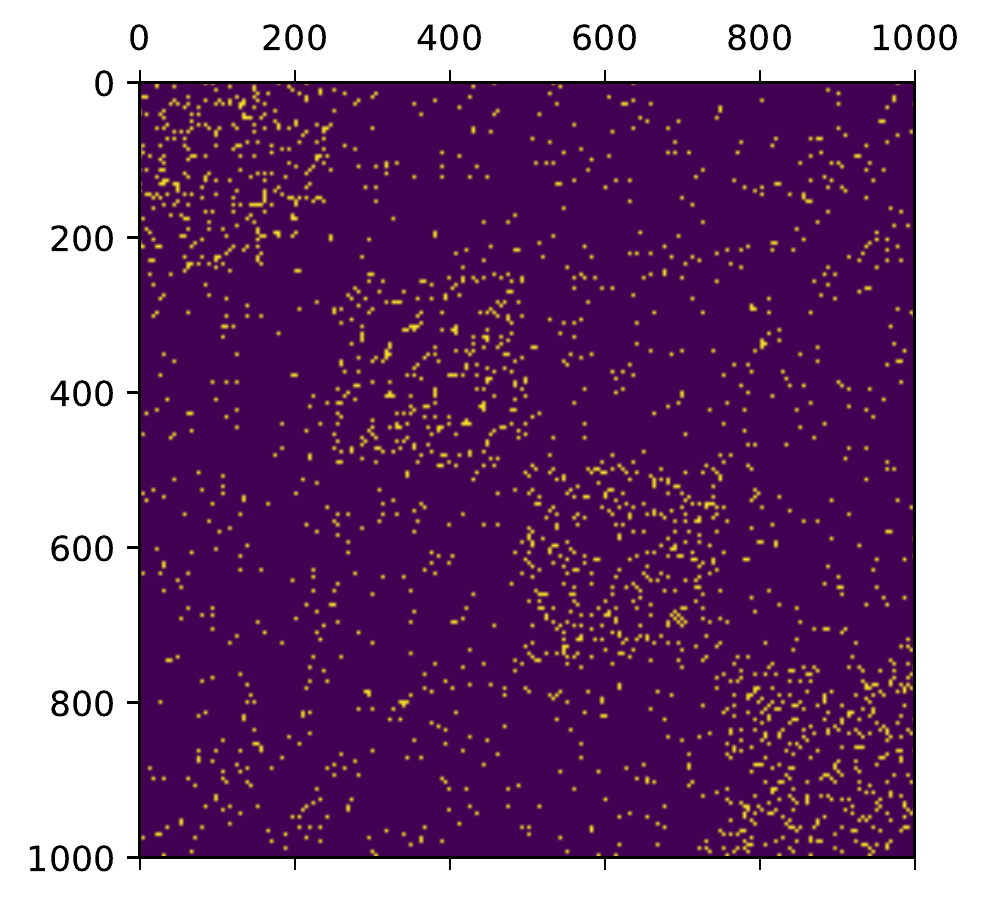} \quad
\includegraphics[width=0.33\linewidth]{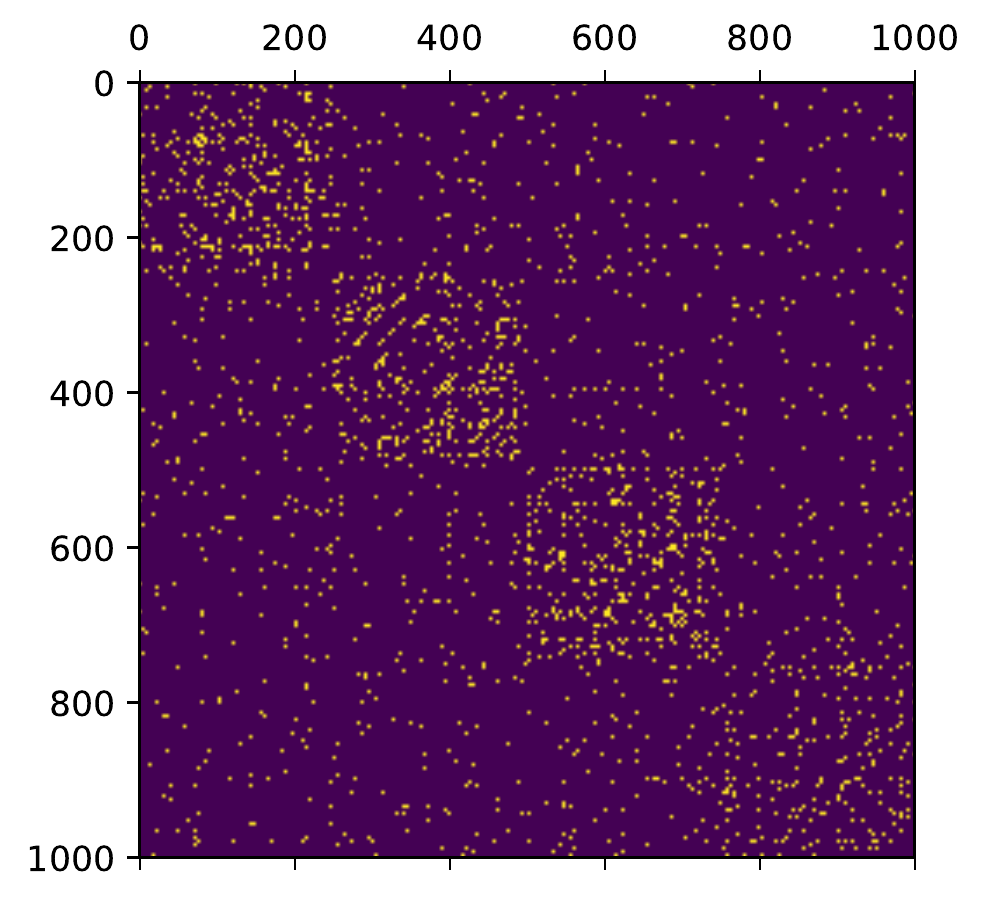}
\caption{$G$ (left), a stochastic block model graph with 1000 nodes and 4 clusters, and $\tilde G$ (right), a reconstruction of $G$ from a $32$-dimensional NetMF embedding. While $G$ and $\tilde G$ differ in the exact edges they contain, we can see that the community structure is preserved.}
\label{fig:intro_matshow}
\end{figure}

\spara{Our contributions.} We make the following findings:  

\begin{itemize}
\item We prove that when the embedding dimension $k$ is equal to $n$ and the node embedding method is NetMF in the limit as the co-occurrence window size parameter goes to infinity, then solving a linear system can provably recover $G$ from $\mathcal{E}(G)$, i.e., find $\tilde G = G$. 

\item We present two algorithms for solving Problem~\ref{prob:inversion} on NetMF embeddings in typical parameter regimes. The first is inspired by the above result, and relies on solving a linear system. 
The second is based on minimizing $\norm{\mathcal{E}(G) - \mathcal{E}(\tilde G)}_F$, where $\norm{\cdot}_F$ is the matrix Frobenius norm, using gradient based optimization.

\item   Despite the non-convex nature of the above optimization problem, we show empirically that our approach successfully solves Problem \ref{prob:inversion} on a variety of real word graphs, for a range of embedding dimensions used frequently in practice.  We show that, typically our optimization based algorithm outperforms  the linear system approach with respect to producing a graph $\tilde G$ with embeddings closer to those of the input graph $G$.

\item We study Problem~\ref{prob:stab} by applying our optimization algorithm to NetMF embeddings for a variety of real world graphs. We compare the input graph $G$ and the output of our inversion algorithm  $\tilde G$ across different criteria.  Our key findings include the following:
\begin{enumerate}
\item  \spara{Fine-Grained Edge Information.} As the embedding dimension $k$ increases {\em up to a certain point}  $\tilde G$ tends closer to $G$, i.e., the Frobenius norm of the difference of the adjacency matrices gets smaller.  After a certain point, the recovery algorithm is trying unsuccessfully to reconstruct fine grained edge information that is ``washed-out'' by NetMF.  
Figure~\ref{fig:intro_fig}(a) illustrates this finding for a popular benchmark of datasets (see Section~\ref{sec:exp} for more details).

\item \spara{Graph properties.} We focus on two fundamental graph properties, counts of triangles and community structure.  Surprisingly, while the number of triangles in $G$ and $\tilde G$ can differ significantly, community structure is well-preserved. In some cases this structure is actually enhanced/emphasized by the embedding method. I.e., the conductance of the same community in $\tilde G$ is even lower than in $G$. 

Figure~\ref{fig:intro_fig}(b) shows the relative error between the conductance of a ground-truth community in $G$ and the conductance of the same community in $\tilde G$ vs. $k$ for the five largest communities in a human protein-protein interaction network.   
\end{enumerate} 

Figure~\ref{fig:intro_matshow} provides another visual summary of the above findings. Specifically, it shows on the left the spy plot of a stochastic block model graph with 1\,000 nodes and four clusters, and on the right the spy plot of the output of our reconstruction algorithm from  a $32$-dimensional NetMF embedding of the former graph. The two graphs differ on exact edges, but the community structure is preserved.
\end{itemize}
 
\section{Related work} 
\label{sec:rel} 
 
\spara{Graph recovery from embeddings.}  To the best of our knowledge, Problem~\ref{prob:inversion} has not been studied explicitly in prior work. \citet{hoskins2018learning} study graph recovery using a partial set of effective resistance measurements between nodes -- equivalent to Euclidean distances for a certain embedding, see Section 4 of \citep{spielman2011graph}. Close to our work lies recent work on node embedding privacy, and in particular graph reconstruction attacks on these embeddings. \citet{EllersCochezSchumacher:2019}  identify  neighbors of a given node $v$ with good accuracy by considering the change in embeddings of the other nodes in $G$ and $G \setminus v$. \citet{DudduBoutetShejwalkar:2020} study a graph reconstruction attack that inverts a simple spectral embedding using a neural network. Training this network requires knowledge of a random subgraph of $G$, used as training data, and can be viewed as solving Problem \ref{prob:inversion}, but with some auxiliary information provided on top of $\mathcal{E}(G)$. 
 
Graph sketching algorithms study the recovery of information about $G$ (e.g., approximations to all its cuts or shortest path distances) from linear measurements of its edge-vertex incidence matrix \citep{McGregor:2014}. These linear measurements can be thought of as low-dimensional node embeddings. However, generally they are designed specifically to encode certain information about $G$, and they differ greatly from the type of embeddings used in graph learning applications. Recently, \citet{chanpuriya2020node} showed that any graph with degree bounded by $\Delta$ admits an embedding into $2\Delta+1$ dimensions that can be \emph{exactly inverted}.  These exact embeddings allow for a perfect encoding of the full graph structure in low-dimensions,  and circumvent limitations of a large family of   embeddings that cannot capture triangle richness and edge sparsity {\em provably} in low dimensions \citep{seshadhri2020impossibility}.

\spara{DeepWalk and NetMF.} We focus on inverting embeddings produced by the \citet{QiuDongMa:2018} NetMF variant of the popular DeepWalk method of \citet{PerozziAl-RfouSkiena:2014}. Consider an undirected, connected, non-bipartite graph $G$, with  adjacency matrix $A \in \{0,1\}^{n \times n}$, diagonal degree matrix $D \in \mathbb{R}^{n \times n}$ and volume $v_G = \tr(D) = \sum_{i, j} A_{i,j}$.
Qui et al. show that, for window size hyperparameter $T$ (typical settings are $T = 10$ or $T = 1$), DeepWalk  stochastically factorizes the \emph{pointwise mutual information (PMI) matrix}:
\begin{align*}
\hat{M}_T = \log \left( \frac{v_G}{T}\sum_{r=1}^{T}(D^{-1}A)^r D^{-1} \right),
\end{align*}

\noindent where the logarithm is applied entrywise to its $n \times n$ argument.  Note that if the diameter of $G$ exceeds $T$, then at least one entry of $\sum_{r=1}^{T}(D^{-1}A)^r D^{-1}$ will be $0$. To avoid taking the logarithm of $0$, NetMF instead employs the \emph{positive pointwise mutual information (PPMI)} matrix:
\begin{align}
\label{eq:netmf}
M_T = \log \left( \max\left(1, \frac{v_G}{T}\sum_{r=1}^{T}(D^{-1}A)^r D^{-1} \right) \right) .
\end{align} 

Via truncated eigendecomposition of $M_T$, one can find an eigenvector matrix $V \in \mathbb{R}^{n \times k}$ and a diagonal eigenvalue matrix $W \in \mathbb{R}^{k \times k}$ such that ${M}_{T,k} = V W V^\top$ is the best possible $k$-rank approximation of $M_T$ in the Frobenius norm. The NetMF embedding is set to the eigenvectors scaled by the square roots of the eigenvalue magnitudes. I.e., $\mathcal{E}(G) = V \sqrt{|W|}$, where the absolute value and the square root are applied entrywise. In practice, these node embeddings perform at least as well as DeepWalk in downstream tasks. Further, their deterministic nature lets us to define a straightforward optimization model to invert them.

\section{Proposed methods}
\label{sec:proposed}

In Sections~\ref{subsec:linear_system} and \ref{subsec:gradalg} we present our two proposed NetMF embedding inversion methods. The first is inspired by our constructive proof of Theorem~\ref{thm:invert} and relies on solving an appropriately defined linear system. The second  is based on optimizing a natural objective using a gradient descent algorithm. Since the NetMF embedding $\mathcal{E}(G)$ encodes the best $k$-rank approximation $M_{T,k} = VWV^T$ to the positive pointwise mutual information (PPMI) matrix $M_T$, we will assume throughout that we are given $M_{T,k}$ directly and seek to recover $\tilde G$ from this matrix. We also assume knowledge of the number of edges in $G$ in terms of the volume $v_G$.

While all networks used in our experiments are unweighted, simple, undirected graphs, i.e., their adjacency matrices are binary ($A \in \{0,1\}^{n \times n}$), our inversion algorithms produce $\tilde G$ with $\tilde{A}\in [0,1]^{n \times n}$. The real valued edge weights in $\tilde G$ can be thought of as representing edge probabilities. We will also convert $\tilde G$ to an unweighted graph with binary adjacency matrix $\tilde{A}_b\in \{0,1\}^{n \times n}$. We describe the binarization process in detail in the following sections. 
 
\subsection{Analytical Approach}
\label{subsec:linear_system}

We leverage a recent asymptotic result of \citet{infiniteWalk}, which shows that as the number of samples and the window size $T$ for DeepWalk/NetMF tend to infinity, the PMI matrix tends to the limit: 
\begin{equation}
\label{eq:infwalk}
\begin{aligned}
\lim_{T\to \infty} T \cdot \hat{M}_T &= \hat{M}_\infty = v_G \cdot D^{-1/2} (\bar L^+ - I) D^{-1/2} + J ,
\end{aligned}
\end{equation}
where $\bar L = I - D^{-1/2} A D^{-1/2}$ is the normalized Laplacian, $\bar{L}^+$ is the Moore-Penrose pseudoinverse of this matrix, and $J$ is the all-ones matrix. Our first observation is that if, in addition to $\hat{M}_\infty$, we are given the degrees of the vertices in $G$, then we know both $D$ and $v_G$, and we can simply invert equation \eqref{eq:infwalk} as follows: 
\begin{equation}
\label{eq:infwalkbackideal}
\begin{aligned}
\bar{L} &=  \left(  D^{1/2} \left(\frac{ \hat{M}_\infty - J}{v_G}\right)  D^{1/2} + I \right)^+ \\
A &=  D^{1/2}  \left( I - \bar{L} \right)  D^{1/2} .
\end{aligned}
\end{equation}  

In Appendix~\ref{app:degrees}, we show using just the graph volume $v_G$, that one can perfectly recover the degree matrix ${D}$ from $\hat{M}_\infty$ via a linear system, provided the adjacency matrix of $G$ is full-rank. 
Combining this fact with Equations~\eqref{eq:infwalk} and~\eqref{eq:infwalkbackideal} we obtain the following: 
 
\begin{theorem}[Limiting Invertibility of Full-Rank PMI Embeddings]
\label{thm:invert}
Let $G$ be an undirected, connected, non-bipartite graph with full-rank adjacency matrix $A \in \{0,1\}^{n \times n}$ and  volume $v_G$. Let $\hat{M}_T$ be the PMI matrix of $G$ which is produced with window size $T$. There exists an algorithm that takes only $\hat{M}_T$ and $v_G$ as input and recovers $A$ exactly in the limit as $T \to \infty$.
\end{theorem}
 
In our embedding inversion task, rather than the exact limiting PMI matrix $\hat{M}_\infty$, we are given the low-rank approximation $M_{T,k}$ of the finite-$T$ PPMI matrix, through the NetMF embeddings. Our first algorithm is based on essentially ignoring this difference. We use $M_{T,k}$ to obtain an approximation to $\hat{M}_\infty$, which we then plug into \eqref{eq:infwalkbackideal}.  This approximation is based on inverting the following limit, shown by~\citet{infiniteWalk}:
\begin{align}\label{eq:pminonlinearapprox}
\lim_{T\to \infty} \hat{M}_T = \log\left( \tfrac{1}{T} \hat{M}_\infty + J \right), 
\end{align}
where the logarithm is applied entrywise. 

Due to the various approximations used, the elements of the reconstructed adjacency matrix  $\tilde A$ may not be in $\{0,1\}$, and may not even be in $[0,1]$; for this reason, as in \citet{seshadhri2020impossibility}, we apply an entrywise clipping function, $\text{clip}(x) = \min(\max(0,x),1)$, after the inversion steps from Equations~\eqref{eq:infwalkbackideal} and \eqref{eq:pminonlinearapprox}. The overall procedure is given in Algorithm~\ref{alg:dwbackanl}.
  
\begin{algorithm}
    \caption{DeepWalking Backwards (Analytical)}
    \label{alg:dwbackanl}
    \textbf{input} approximation $M_{T,k}$ of true $T$-step PPMI, window-size $T$, degree matrix $D$, graph volume $v_G$ \\
    \textbf{output} reconstructed adjacency matrix $\tilde{A} \in [0,1]^{n \times n}$
    \begin{algorithmic}[1] 
	\State $\tilde{M}_\infty \gets T \cdot \left( \exp \left({M_{T,k}} \right) - J \right)$ \Comment{\textcolor{gray}{$\exp$ is applied entrywise, $J$ is the all-ones matrix}}
	\State $\tilde{\bar{L}} \gets  \left( D^{1/2} \left(\frac{ \tilde{M}_\infty - J}{v_G}\right)  D^{1/2} + I \right)^+$
	\State $\tilde{A} \gets \text{clip}\left(  D^{1/2}  \left( I - \tilde{\bar{L}} \right)  D^{1/2} \right)$
            \State \textbf{return} $\tilde{A}$
    \end{algorithmic}
\end{algorithm}

\spara{Binarization.} To produce a binary adjacency matrix $\tilde{A}_b\in \{0,1\}^{n \times n}$ from $\tilde{A}$, we use a slight modification of Algorithm~\ref{alg:dwbackanl}: rather than clipping, we set the highest $v_G$ off-diagonal entries above the diagonal to 1, and their symmetric counterparts below the diagonal to 1. This ensures that the matrix represents an undirected graph $\tilde G$ with the same number of edges as $G$. 

\subsection{Optimization Approach}
\label{subsec:gradalg} 

Our gradient based approach parameterizes the entries of a real valued  adjacency matrix $\tilde{A} \in (0,1)^{n \times n}$ with independent logits for each potential edge, and leverages the differentiability of Equation~\eqref{eq:netmf}. Based on $\tilde{A}$, we compute the PPMI matrix $\tilde{M}_T$, and then the squared PPMI error loss, i.e., the squared Frobenius error between $\tilde{M}_T$ and the low-rank approximation $M_{T,k}$ of the true PPMI, given by the NetMF embeddings. We differentiate through these steps, update the logits, and repeat. Pseudocode is given in Algorithm~\ref{alg:dwbackgrad}.

Since the input to the algorithm is a low-rank approximation of the true PPMI, and since this approximation is used for the computation of error, it may seem more appropriate to also compute a low-rank approximation of the reconstructed PPMI matrix $\tilde M_T$ prior to computing the error; we skip this step since eigendecomposition within the optimization loop is both computationally costly and unstable to differentiate through.

Note that we invoke a ``shifted logistic'' function $\sigma_v$ which constructs an adjacency matrix with a given target volume. The pseudocode for this function is given in Algorithm~\ref{alg:slogistic}. This algorithm is an application of Newton's method. We find that 10 iterations are sufficient for convergence in our experiments.

Our implementation uses PyTorch~\citep{NEURIPS2019_9015} for automatic differentiation and minimizes the loss using the SciPy \citep{scipy} implementation of L-BFGS \citep{liu1989limited,zhu1997algorithm} with default hyperparameters and a maximum of 500 iterations.

\begin{algorithm}
    \caption{DeepWalking Backwards (Optimization)}
    \label{alg:dwbackgrad}
    \textbf{input} approximation $M_{T,k}$ of true $T$-step PPMI, window-size $T$, graph volume $v_G$, number of iters. $N$ \\
    \textbf{output} reconstructed adjacency matrix $\tilde{A} \in (0,1)^{n \times n}$
    \begin{algorithmic}[1] 
	\State Initialize elements of $X \in \mathbb{R}^{(n \times n)}$ to $0$ \Comment{\textcolor{gray}{logits of the reconstructed adjacency matrix}}
	\For{$i \gets 1$ to $N$}                    	       
		\State $\tilde{A} \gets \sigma_{v_G}(X)$ \Comment{\textcolor{gray}{construct adjacency matrix with target volume, see Algorithm~\ref{alg:slogistic}}}
		\State $\tilde{M}_T \gets \text{PPMI}\left( \tilde{A} \right)$ via Eq.~\eqref{eq:netmf}
		\State $L \gets \Vert \tilde{M}_T - M_{T,k} \Vert_F^2 $ \Comment{\textcolor{gray}{squared error of PPMI}}
		\State Calculate $\partial_{X} L$ via automatic differentiation through Steps 3 to 5
		\State Update $X$ to minimize $L$ using $\partial_{X} L$
	\EndFor
            \State \textbf{return} $\sigma_v(X)$
    \end{algorithmic}
\end{algorithm}

\begin{algorithm}
    \caption{Shifted Logistic Function $\sigma_v$}
    \label{alg:slogistic}
    \textbf{input} logit matrix $X \in \mathbb{R}^{(n\times n)}$, target sum $v \in (0,n^2)$, number of iterations $I$ \\
    \textbf{output} matrix $A \in (0,1)^{n \times n}$ which sums approximately to $v$
    \begin{algorithmic}[1] 
        \State $s \gets 0$
        \For{$i \gets 1$ to $I$}
                \State $A \gets \sigma(X+s)$ \Comment{\textcolor{gray}{ $\sigma$ is the logistic function applied entrywise }}
                \State $s \gets s + \frac{ v - \Sigma(A) }{\Sigma\left( A \circ (1 - A) \right)}$ \Comment{\textcolor{gray}{ $\Sigma$ sums over all elements, and $\circ$ is an entrywise product }}
        \EndFor
           \State \textbf{return} $\sigma(X+s)$
    \end{algorithmic}
\end{algorithm}

\spara{Binarization.} We binarize the reconstructed $\tilde{A} \in (0,1)^{n \times n}$ differently from the prior approach. We treat each element of $\tilde{A}$ as the parameter of a Bernoulli distribution and sample independently to produce $\tilde{A}_b \in \{0,1\}^{n \times n}$. Since we set $\tilde{A}$'s volume to be approximately $v_G$ using the $\sigma_v$ function, the number of edges in the binarized network after sampling is also $\approx v_G$.

\section{Experimental results}
\label{sec:exp}

\subsection{Experimental setup}

\spara{Datasets.}  We apply the NetMF inversion algorithms described in Section~\ref{sec:proposed} to a benchmark of networks, summarized in Table~\ref{tab:datasets}. As part of our investigation of how well the output $\tilde{G}$ of our methods matches the underlying graph $G$, we examine how community structure is preserved. For this reason, we choose only test graphs with labeled ground-truth communities. All datasets we use are publicly available: see \citet{QiuDongMa:2018} for \textsc{BlogCatalog} and \textsc{PPI}, \citet{sen2008collective} for \textsc{Citeseer} and \textsc{Cora}, and SNAP~\citep{snapnets} for \textsc{Email} and \textsc{Youtube}. The \textsc{YouTube} graph we use is a sample of 20 communities from the raw network of \cite{snapnets}. For all networks, we consider only the largest connected component. The community labels that we report for various datasets, such as those reported in the legends of Figure~\ref{fig:conductancesmall}, refer to the labels as given in the input datasets.
%
%
%
%
\begin{table}[h]
\small
\centering
\begin{tabular}{llll}
        \toprule
        \textbf{Name} & \textbf{Nodes} & \textbf{Edges} & \textbf{\# Labels} \\
        \midrule
        {\sc BlogCatalog}   & 10,312 & 333,983 & 39 \\
        {\sc E-mail}  & 986 & 16,064 & 42 \\
        {\sc PPI}   & 3,852 & 76,546 & 50 \\
        {\sc Cora}  & 2,485 & 10,138 & 7 \\
        {\sc Citeseer}  & 2,110 & 7,388 & 6 \\
	{\sc YouTube}  & 10,617 & 55,864 & 20 \\
        \bottomrule
\end{tabular}
\caption{\label{tab:datasets} Datasets used in our experiments.}
\end{table}

\spara{Hyperparameter settings.} We experiment  with  a set of different values  for the embedding dimension $k$, starting from $2^4$ and incrementing in powers of $2$, up to $2^{11}=2048$, except for the {\sc Email} dataset, which has fewer than $2^{10}$ nodes.  For this dataset we only test for $k$ up to $2^9$. 
Throughout the experiments, we set the window-size $T$ to 10, as this is the most commonly used value in downstream machine learning tasks.  

\spara{Evaluation.} Our first step is to evaluate how well the two algorithms proposed in Section \ref{sec:proposed} solve embedding inversion (Problem \ref{prob:inversion}). To do this, 
we measure the error in terms of the relative Frobenius error between the rank-$k$ approximations of the true and reconstructed PPMI matrices, $M_{T,k}$ and $\tilde M_{T,k}$ respectively. These matrices represent the NetMF embeddings of $G$ and $\tilde G$. 
The relative Frobenius error for two matrices $X$ and $\tilde{X}$ is simply $\norm{X-\tilde X}_F/\norm{X}_F$.

We next study how the reconstructed graph $\tilde G$ obtained via embedding inversion compares with the true $G$ (Problem \ref{prob:stab}). Here, we binarize the reconstructed adjacency matrix to produce $\tilde A_b$. See Sections \ref{subsec:linear_system} and \ref{subsec:gradalg} for details. Thus, like $G$, $\tilde G$ is an undirected, unweighted graph.
Most directly, we measure the relative Frobenius error between $G$'s adjacency matrix $A$ and $\tilde G$'s adjacency matrix $\tilde A_b$.   
We also measure the reconstruction error for three other key measures:  
 
\begin{itemize}
\item \textbf{Number of triangles ($\tau$)}. The total number of 3-cliques, i.e., triangles, in the graph. 
\item \textbf{Average path length ($\ell$)}. The average path length between any two nodes in the graph.
\item \textbf{Conductance ($\phi$) of ground-truth communities}. For a community $S$, the conductance is defined as: $\phi(S) = \frac{e(S:\bar{S})}{\min(\text{vol}(S),\text{vol}(\bar{S}))}$
where $e(S:\bar{S})$ is the number of edges leaving community $S$ and $\text{vol}(S)$ is number of edges induced by $S$. $\bar S$ is the complement $V \setminus S$.
\end{itemize}

For the above measures we report the {relative error} between the measure $x$ for the true network and the one of the recovered network $\tilde x$, defined as $(\tilde x-x)/x$. 

Finally, we evaluate how well $\tilde G$'s low-dimensional embeddings perform in classification, where the goal is to infer the labels of the nodes of $G$.
We train a linear model using a fraction of the labeled nodes of $G$ and the low-dimensional embedding of $\tilde G$, and try to infer the labels
of the remaining nodes. We report accuracy in terms of micro F1 score and compare it with the accuracy when using the low-dimensional embedding of $G$ itself. For this task, we use both the recovered real-valued adjacency matrix of $\tilde G$ and its binarized version. We observe that, contrary to the previous measures,
performance is sensitive to binarization.

\spara{Code.} All code is written in Python and is available at \url{https://github.com/konsotirop/Invert_Embeddings}. 

\begin{figure}
    \centering 
    \includegraphics[width=0.40\linewidth]{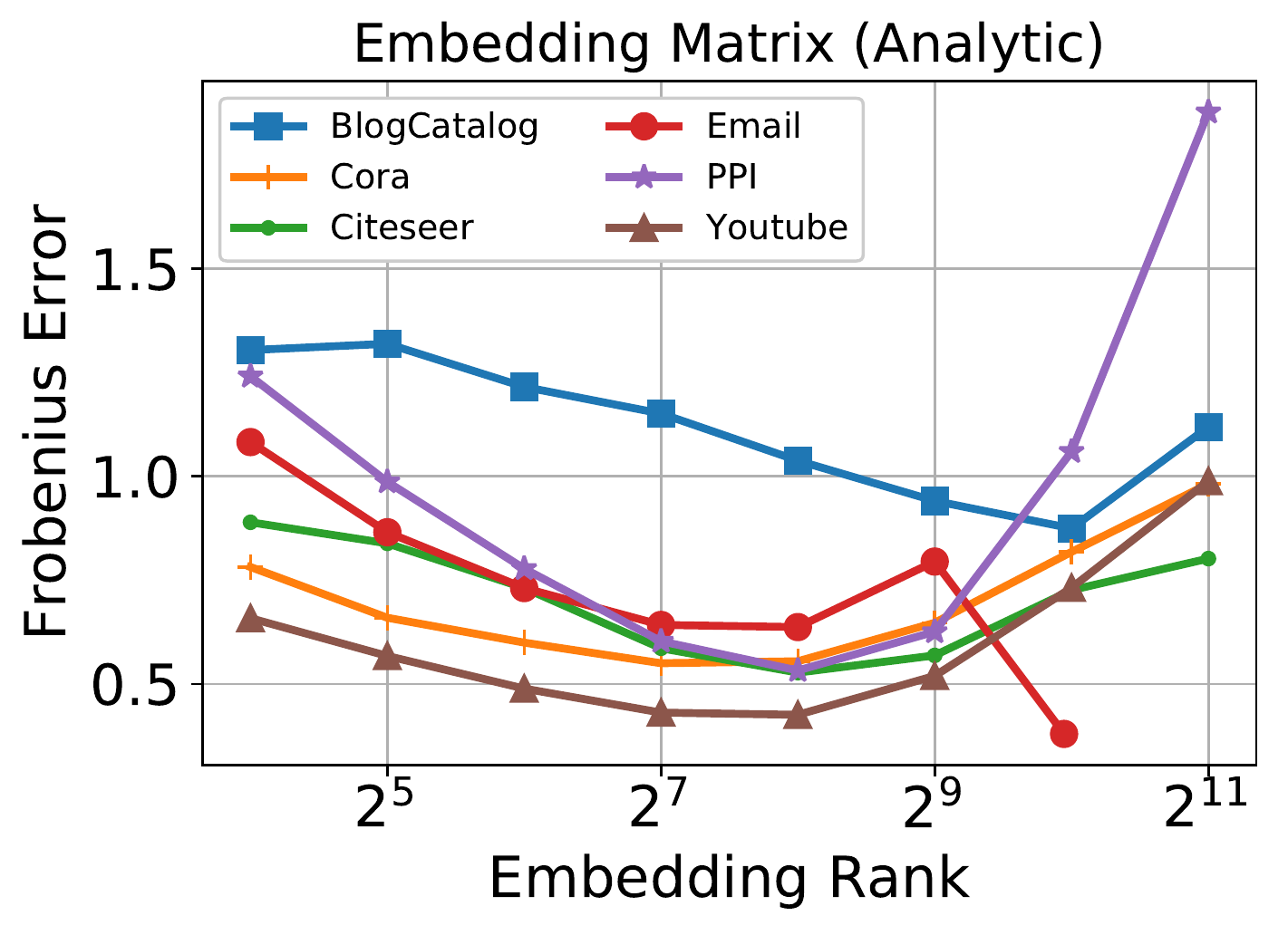} \quad  
     \includegraphics[width=0.40\linewidth]{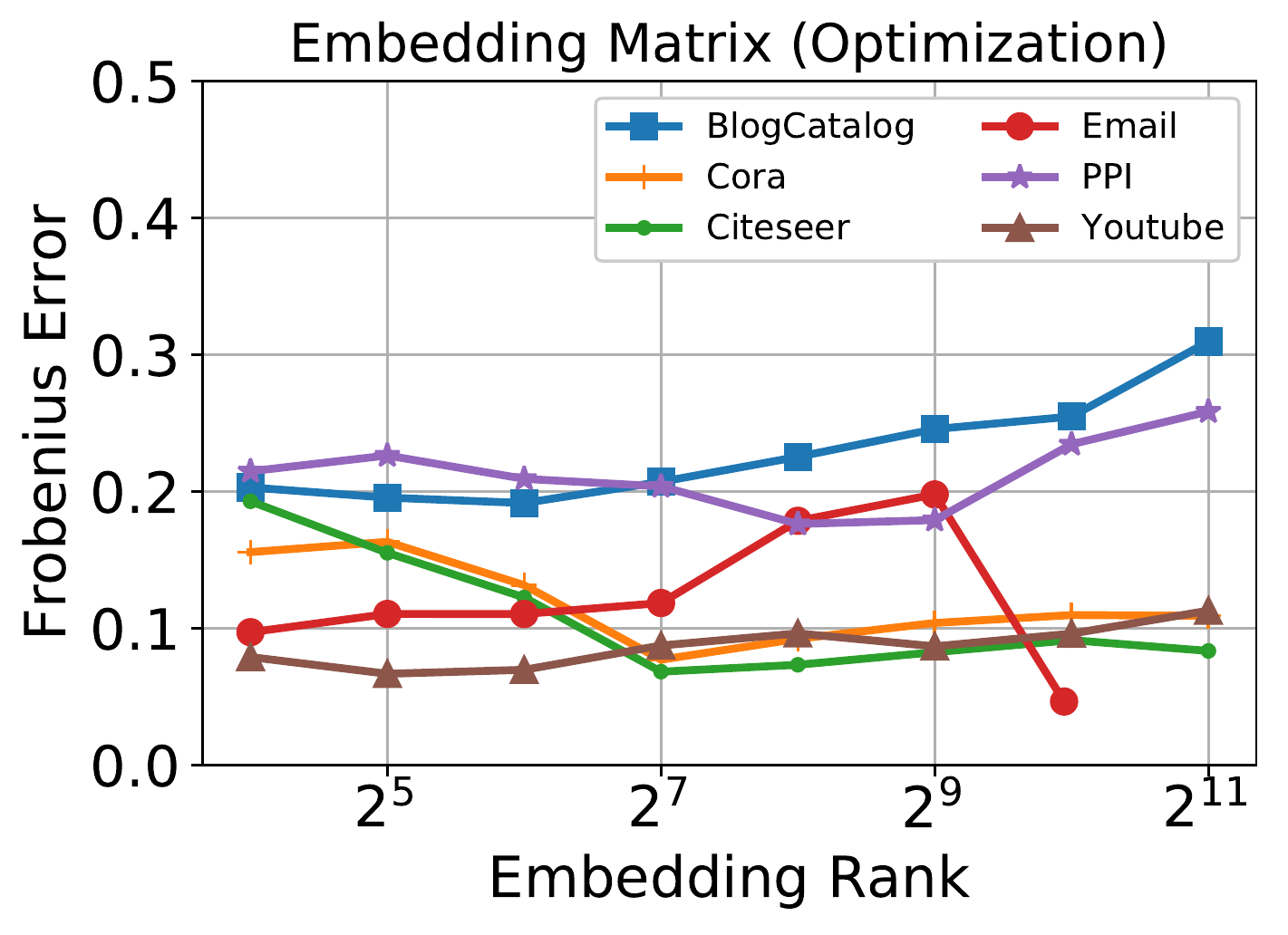} 
    \caption{ \label{fig:frobenius_embedding}  Relative Frobenius error vs. embedding rank $k$ for the low-rank PPMI matrices of the graphs reconstructed using the inversion algorithms: the analytical approach, Alg.~\ref{alg:dwbackanl} (left), and the optimization approach, Alg.~\ref{alg:dwbackgrad} (right). For details, see Section~\ref{subsec:alg1alg2}.
    }
    \vspace{-1.5em}
\end{figure}

\spara{Summary of findings.} Before we delve into details, we summarize our key findings. 

\begin{figure*}
\centering
\includegraphics[width=0.32\linewidth]{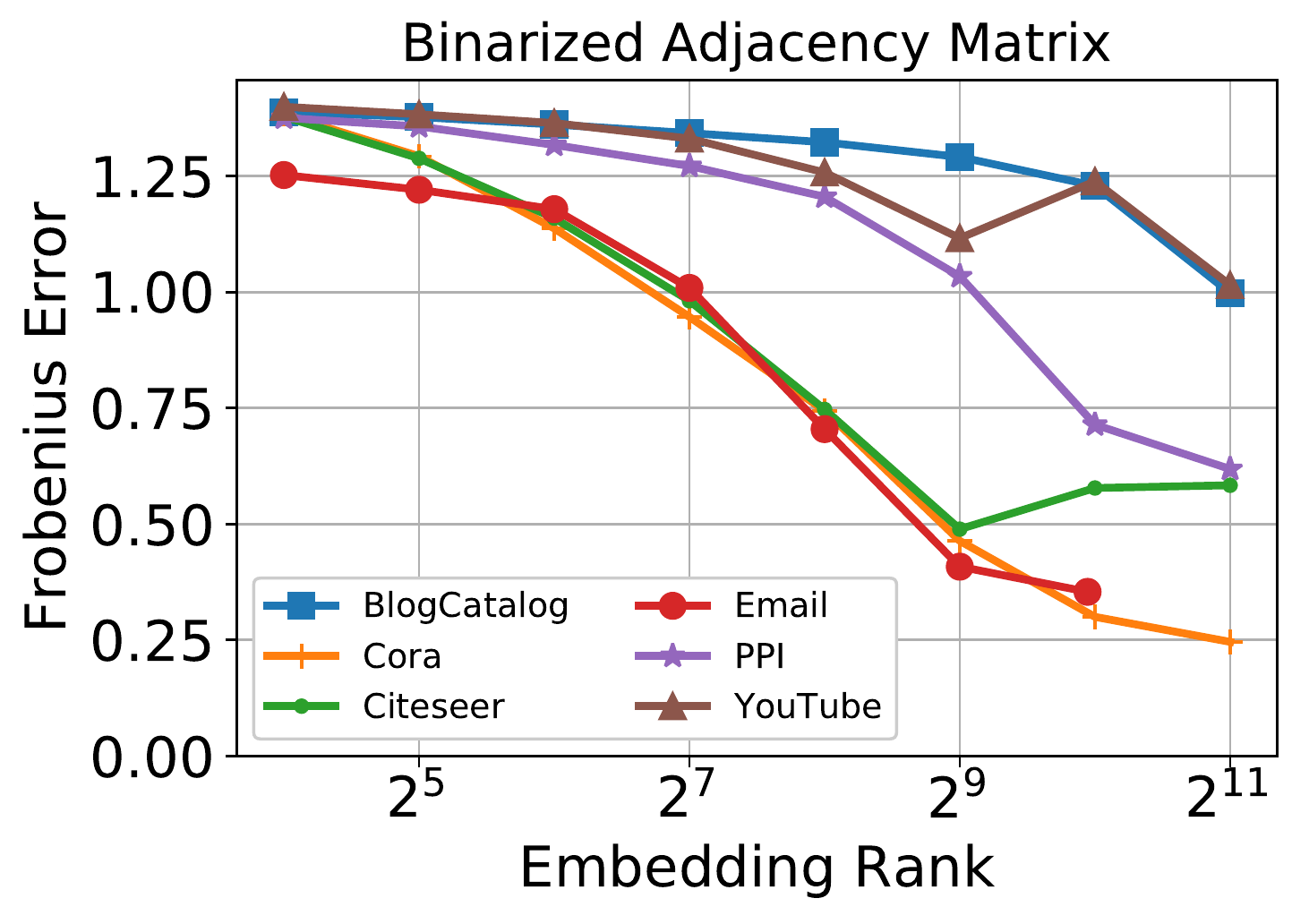} \hfill
\includegraphics[width=0.32\linewidth]{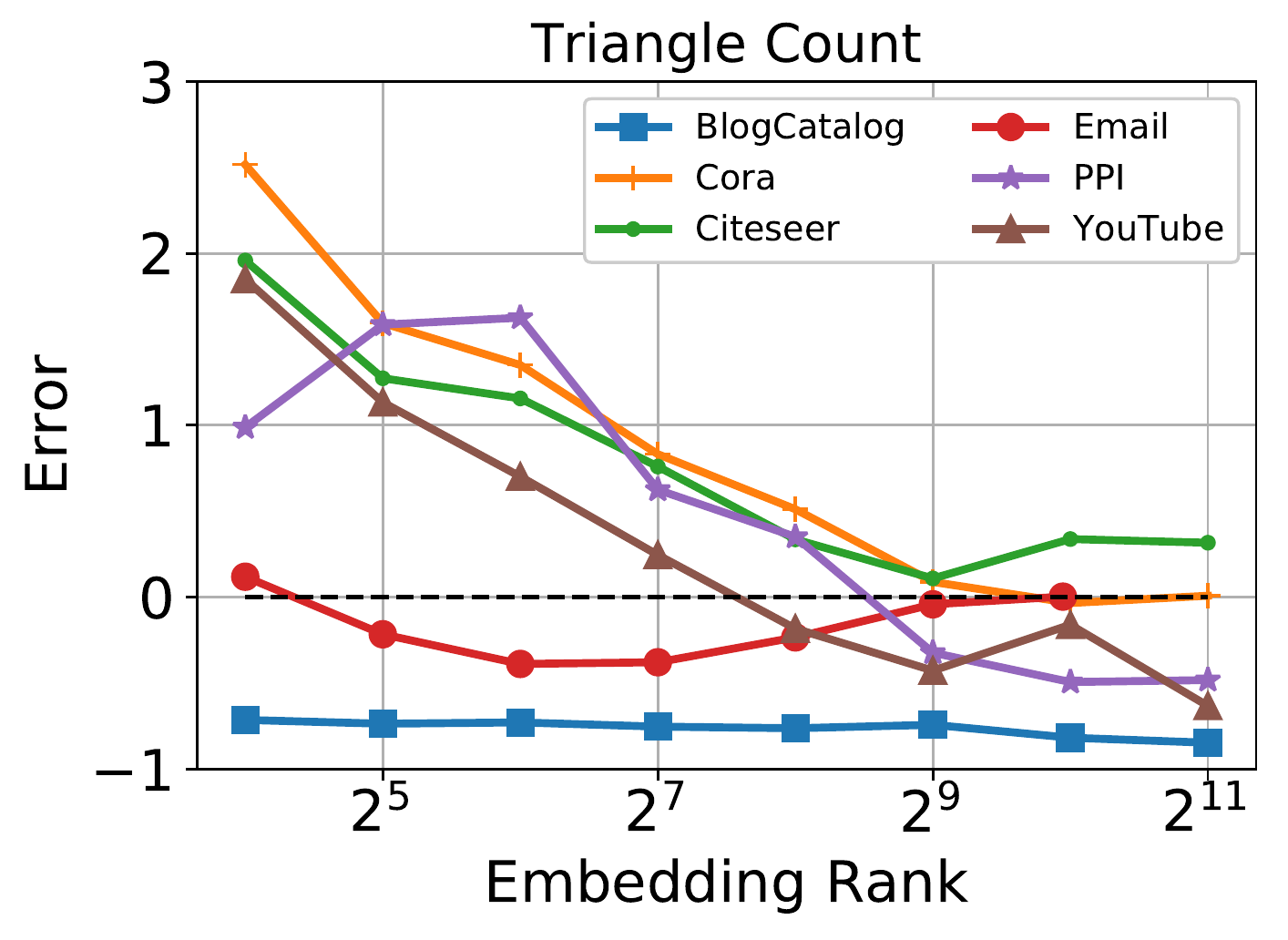}\hfill
\includegraphics[width=0.32\linewidth]{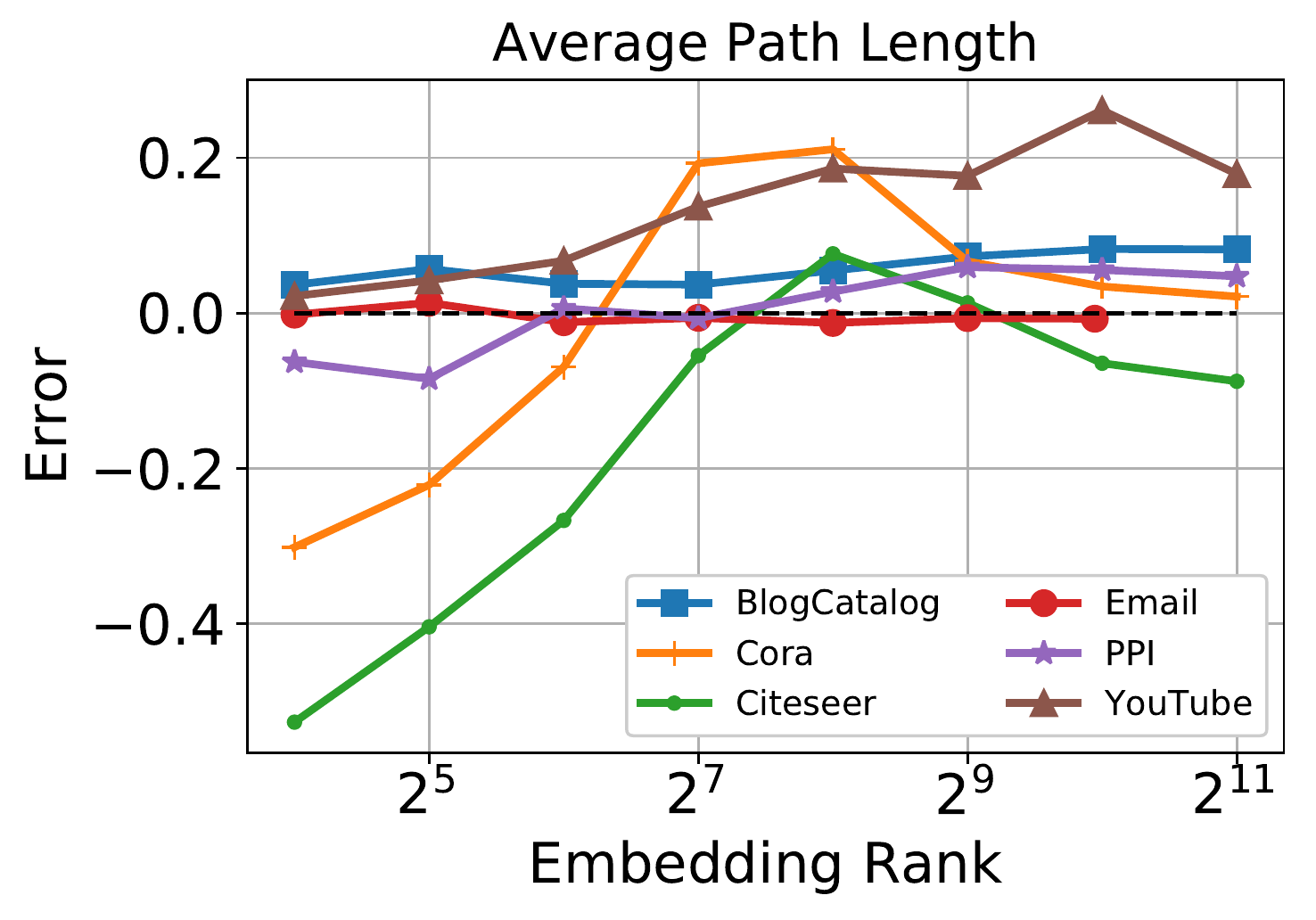}
\caption{\label{fig:alg2_evaluation}   From left to right: Relative Frobenius error for the binarized adjacency matrix; relative error for the number of triangles; and relative error for the average path length. All plots are versus the embedding rank.  
}
\end{figure*}

\begin{itemize}
\item The optimization approach (Alg.~\ref{alg:dwbackgrad}), significantly outperforms the analytical approach (Alg.~\ref{alg:dwbackanl}), in terms of how closely the NetMF embeddings of the reconstructed graph $\tilde G$ match those of the true graph $G$ (i.e., in solving Problem \ref{prob:inversion}). See Figure~\ref{fig:frobenius_embedding}.

\item Focusing on $\tilde G$ produced by Algorithm~\ref{alg:dwbackgrad}, the NetMF embedding is close to the input at all ranks. The adjacency matrix error of $\tilde G$ trends downwards as the embedding rank $k$ increases. However, for small $k$, the two graph topologies can be very different in terms of edges and non-edges. See Figure~\ref{fig:alg2_evaluation}.

\item $\tilde G$ preserves and or even enhances the community structure present in $G$, and tends to preserve the average path length. However, the number of triangles in $\tilde G$ greatly differs from that in $G$ when the embedding rank $k$ is low. See Figure \ref{fig:alg2_evaluation}.

\item $\tilde G$'s NetMF embeddings perform essentially identically to $G$'s in downstream classification on $G$. However, binarization has a significant effect: if we first binarize $\tilde G$'s edge weights, and then produce embeddings, there is a drop in classification performance.

\item Overall, we are able to invert NetMF embeddings as laid out in Problem~\ref{prob:inversion} and, in the process, recover $\tilde G$ with similar community structure to the true graph $G$. Surprisingly, however, $\tilde G$ and $G$ can be very different graphs in terms of both specific edges and broader network properties, despite their similar embeddings. 
\end{itemize}

\subsection{Analytical vs. Optimization Based Inversion}
\label{subsec:alg1alg2}

Figure~\ref{fig:frobenius_embedding} reports the relative Frobenius error of the analytical method (Alg.~\ref{alg:dwbackanl}) and the optimization approach (Alg.~\ref{alg:dwbackgrad}) in embedding inversion as we range $k$.  We can see that   Alg.~\ref{alg:dwbackgrad} {\em significantly} outperforms Alg.~\ref{alg:dwbackanl}.  
While Alg.~\ref{alg:dwbackanl} comes with strong theoretical guarantees (Theorem~\ref{thm:invert}) in asymptotic settings  (i.e., $T \rightarrow \infty$, $k=n$), it performs poorly when these conditions are violated.  
In practice, the embedding dimension $k$ is always set to be less than $n$ (typical values are $128$ or $256$), and $T$ is finite ($T$ is often set to $10$). At these settings, the approximations used in Alg.~\ref{alg:dwbackanl} seem to severely limit its performance.

Given the above, in the following sections we focus our attention on the optimization approach. This approach makes no assumption on the rank $k$, or the window-size $T$. We can see in Figure \ref{fig:frobenius_embedding} that the embedding error stays low across different values of $k$ when using Alg.~\ref{alg:dwbackgrad}, indicating that performance is insensitive to the dimension parameter.

\subsection{Evaluating Graph Recovery} 
\label{subsec:evaluatealg2} 
 \spara{Adjacency matrix reconstuction.} We next examine how closely the output  of Alg.~\ref{alg:dwbackgrad}, the binarized adjacency matrix $\tilde A_{b}$, matches the original adjacency matrix $A$, especially as we vary the embedding dimensionality $k$.  As can be seen in  Figure~\ref{fig:alg2_evaluation}, at low ranks, the relative Frobenius error is often quite high -- near $1$. In combination with Figure~\ref{fig:frobenius_embedding} (left), this shows an interesting finding: two graphs may be very different topologically, but still have very similar low-dimensional node embeddings (i.e., low-rank PPMI matrices). We do observe that as the embedding dimension grows, the adjacency matrix error decreases. This aligns with the message of Theorem~\ref{thm:invert} that, in theory, high dimensional node embeddings yield enough information to facilitate full recovery of the underlying graph $G$.
 We remark that, by construction, $G$ and $\tilde G$ have approximately the same number of edges. Thus, the incurred Frobenius error is purely due to a reorientation of the specific edges between the true and the reconstructed networks.

 \spara{Recovery of graph properties.}    Bearing in mind that the recovered $\tilde G$ differs substantially from the input graph $G$ in the specific edges it contains, we next investigate whether the embedding inversion process at least recovers bulk graph properties.
 
 Figure~\ref{fig:alg2_evaluation} shows the relative error of the triangle count versus embedding dimensionality $k$. We observe that the number of triangles can be hugely different among the true and the reconstructed networks when $k$ is small. In other words, there exist networks with similar low-dimensional NetMF embeddings that differ significantly in their total number of triangles. This is surprising: since the number of triangles is an important measure of local connectivity, one might expect it to be preserved by the node embeddings. In constrast, for another important global property, the average path length, the reconstruction error is always relatively low (also shown in Figure~\ref{fig:alg2_evaluation}).

\begin{figure}
\centering
\includegraphics[width=0.32\linewidth]{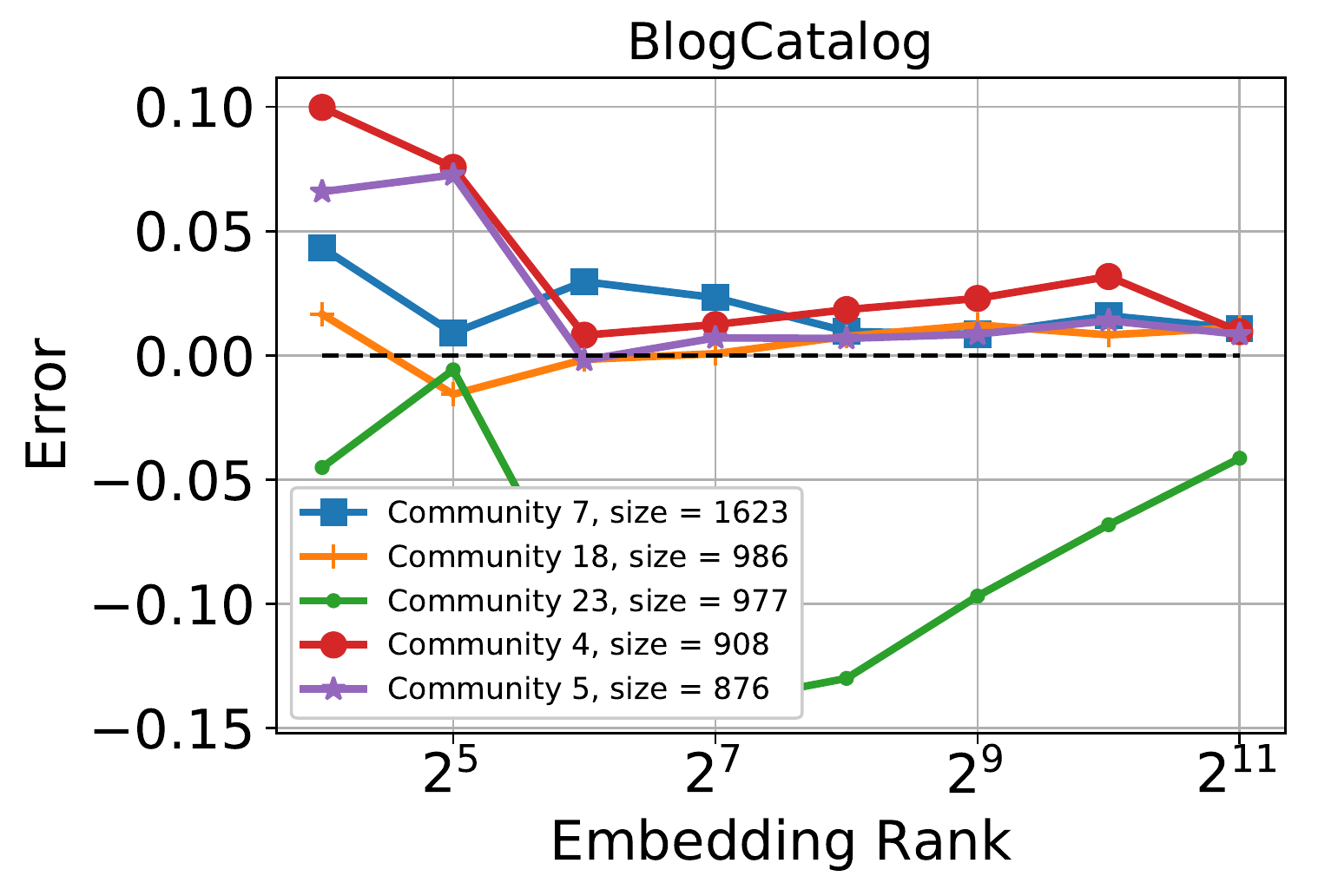} \hfill
\includegraphics[width=0.32\linewidth]{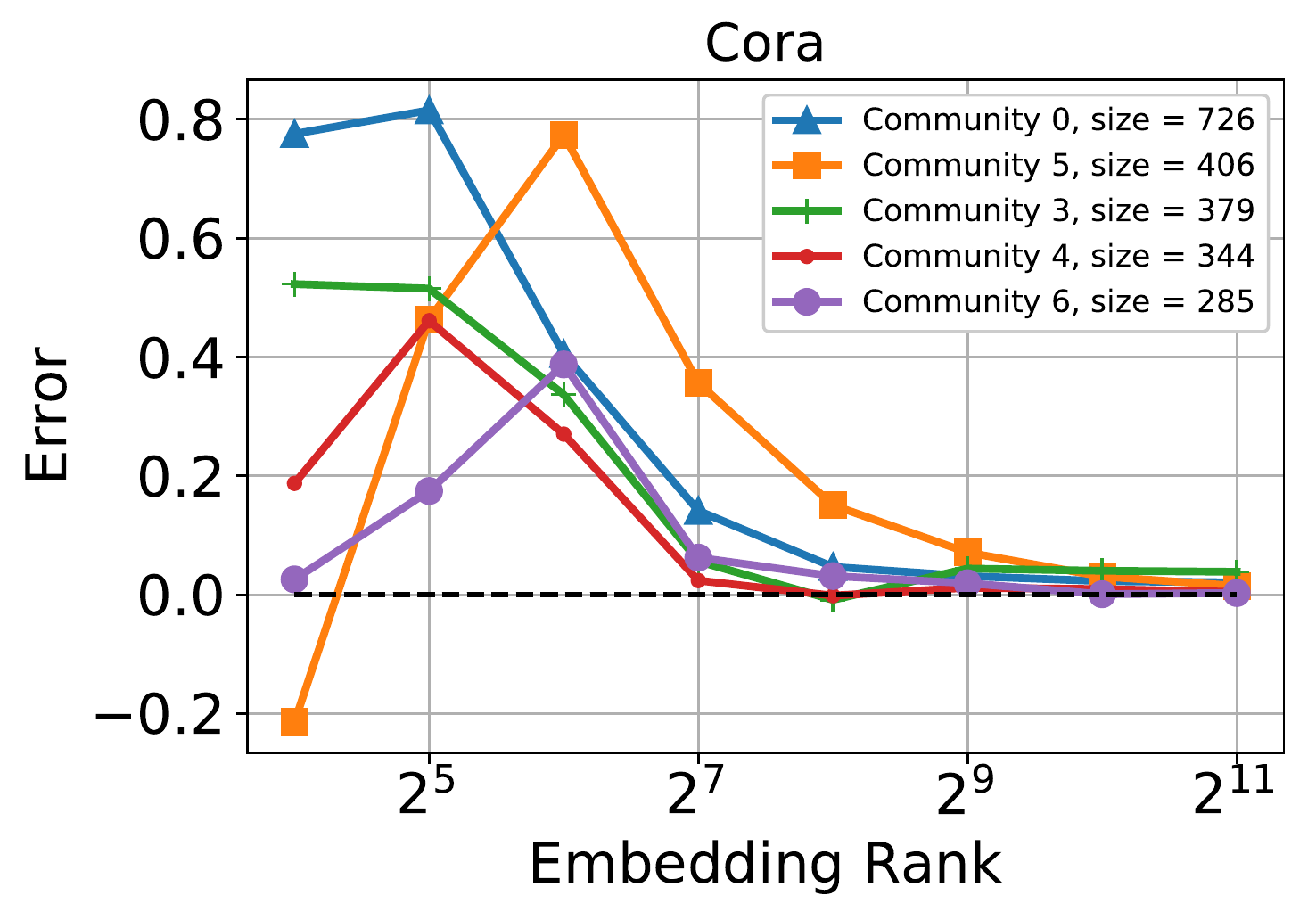} \hfill
\includegraphics[width=0.32\linewidth]{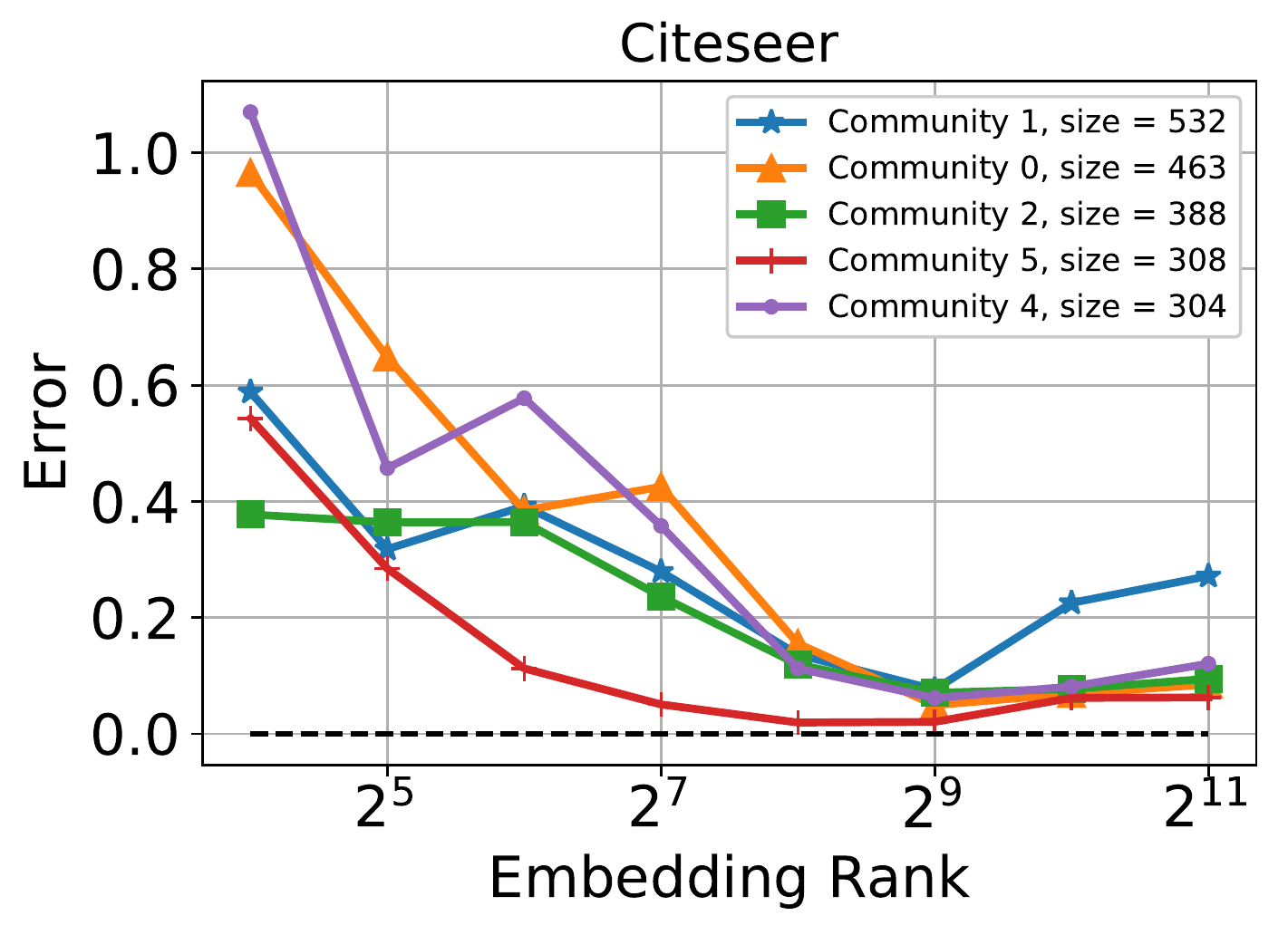}\\
\includegraphics[width=0.32\linewidth]{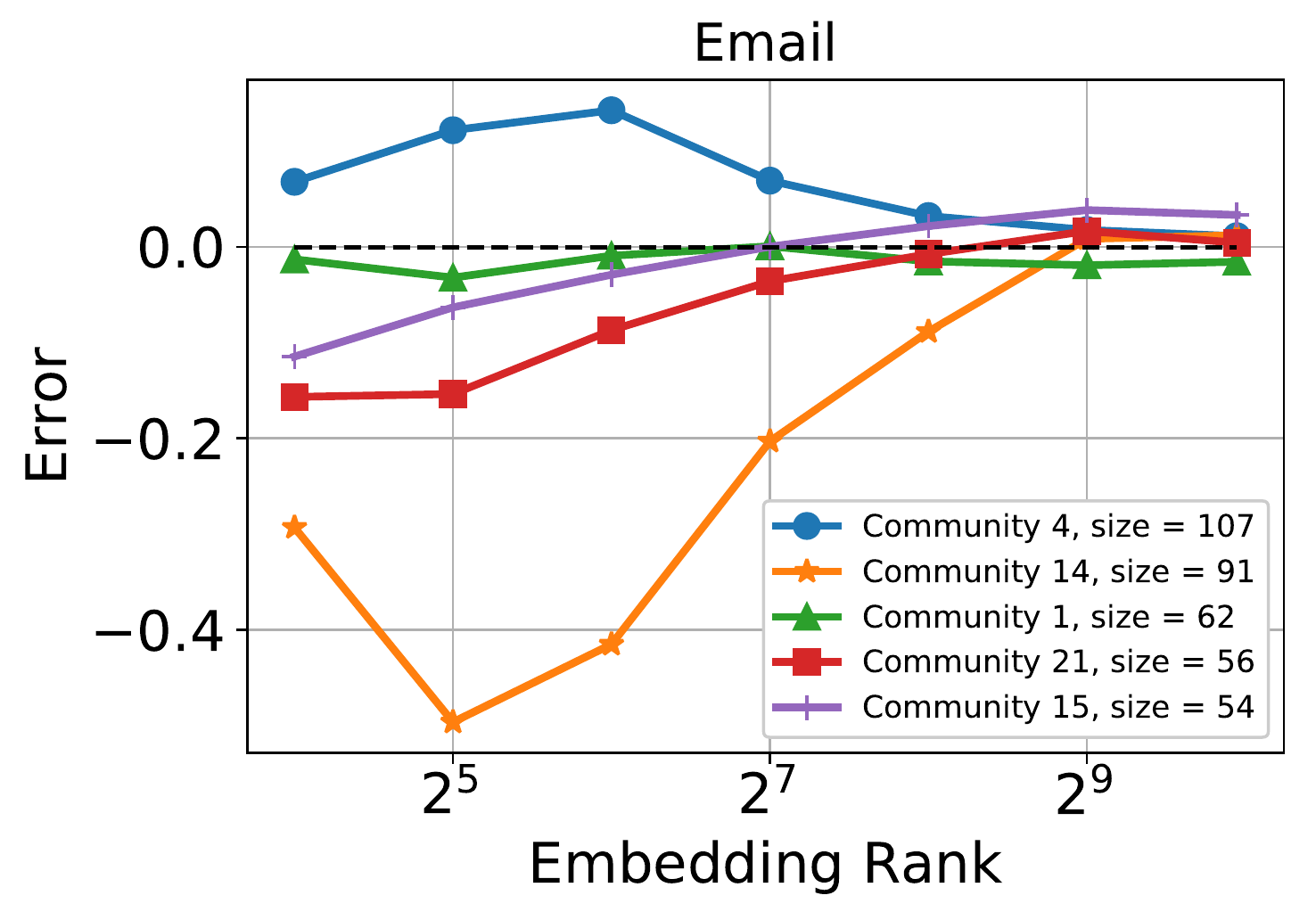} \hfill
\includegraphics[width=0.32\linewidth]{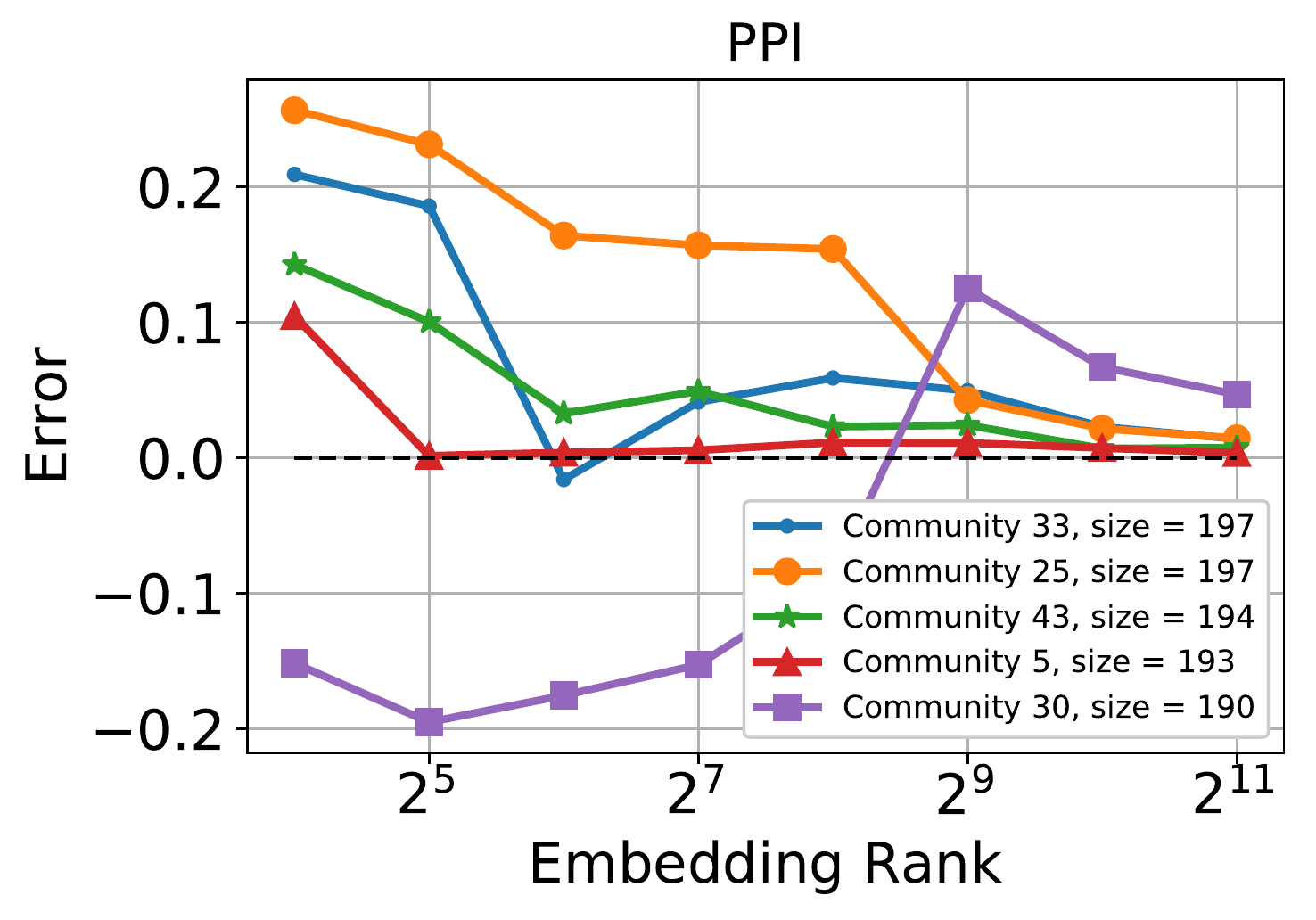}\hfill
\includegraphics[width=0.32\linewidth]{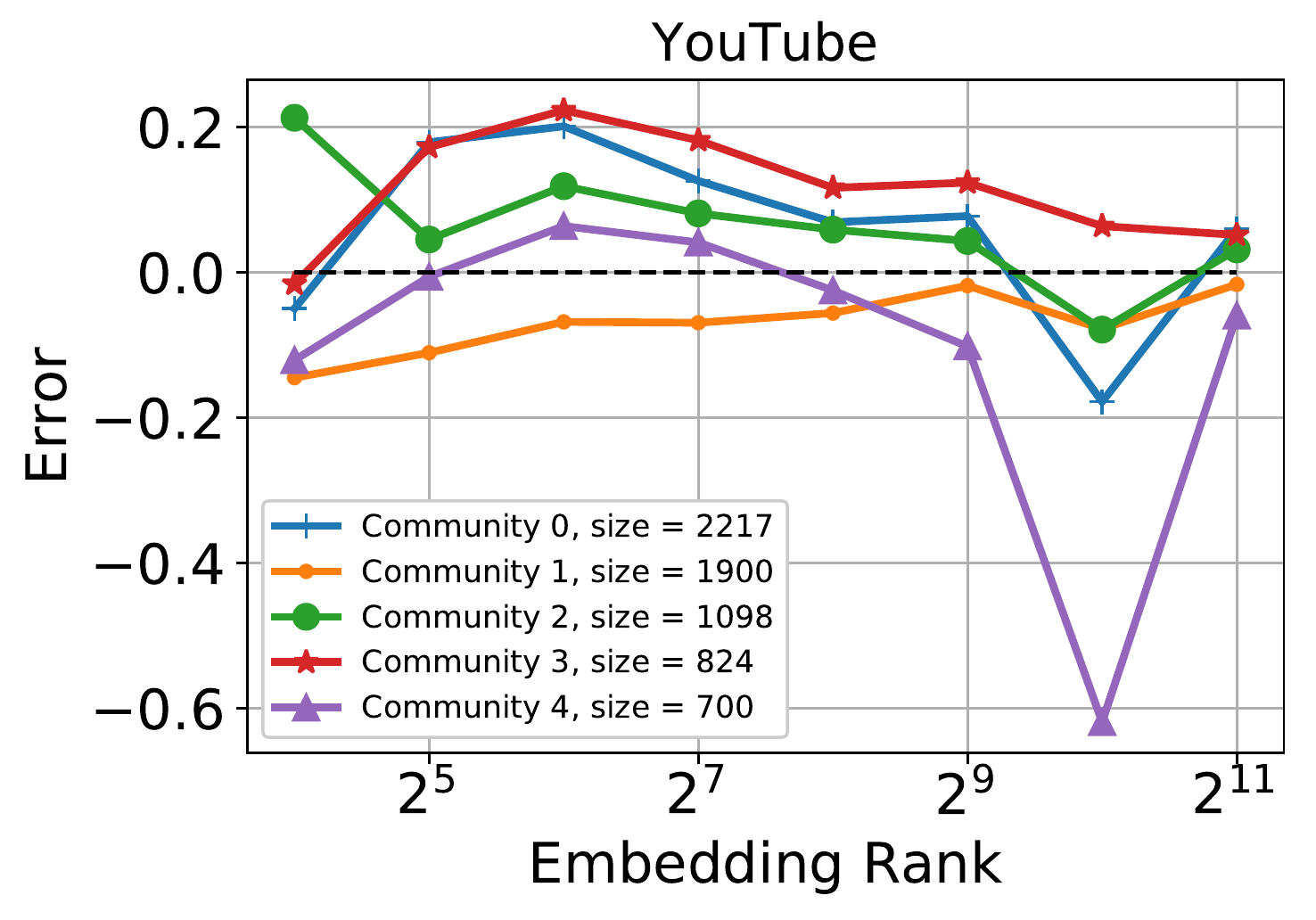}
\caption{Relative error for the conductances of the five largest communities for each of the selected networks.}
\label{fig:conductancesmall}
\end{figure}

\begin{figure}
\centering
\includegraphics[width=0.32\linewidth]{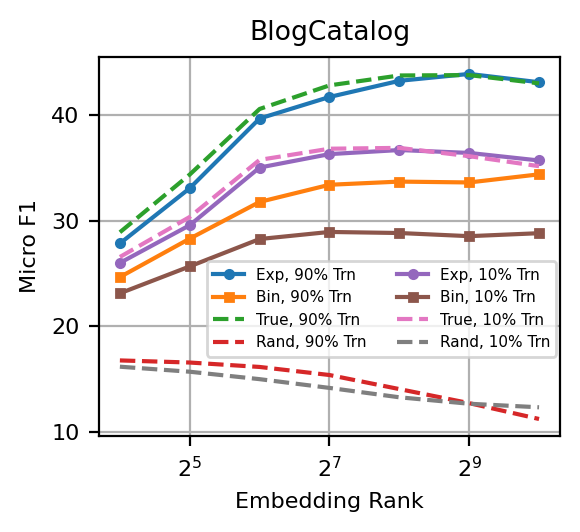} \hfill
\includegraphics[width=0.32\linewidth]{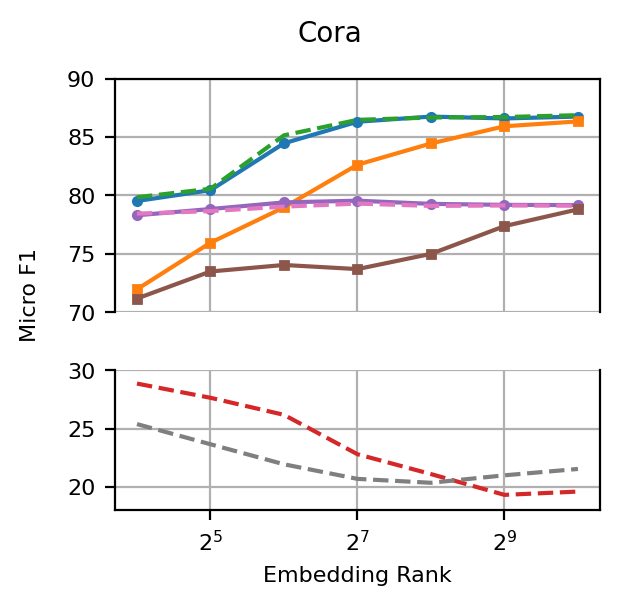}\hfill
\includegraphics[width=0.32\linewidth]{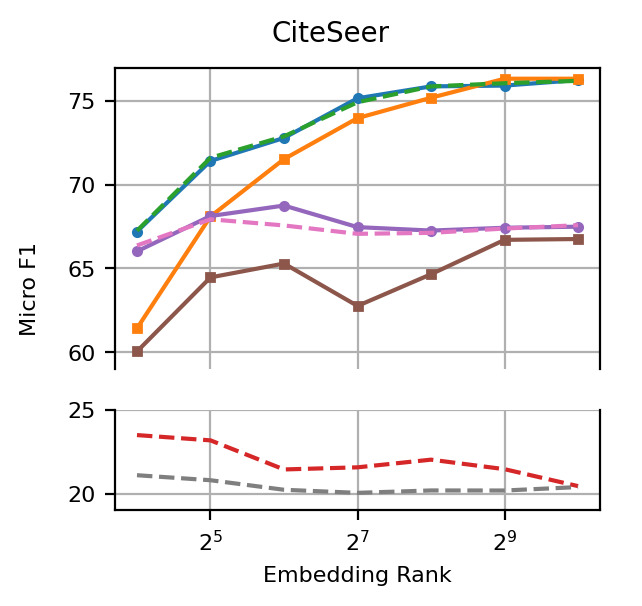} \\ 
\includegraphics[width=0.32\linewidth]{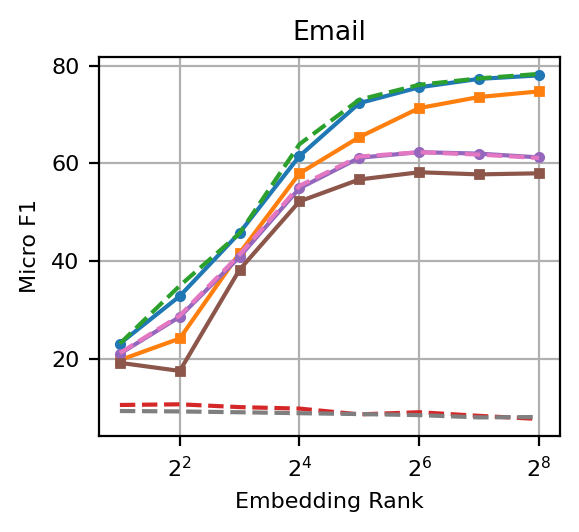}\hfill
\includegraphics[width=0.32\linewidth]{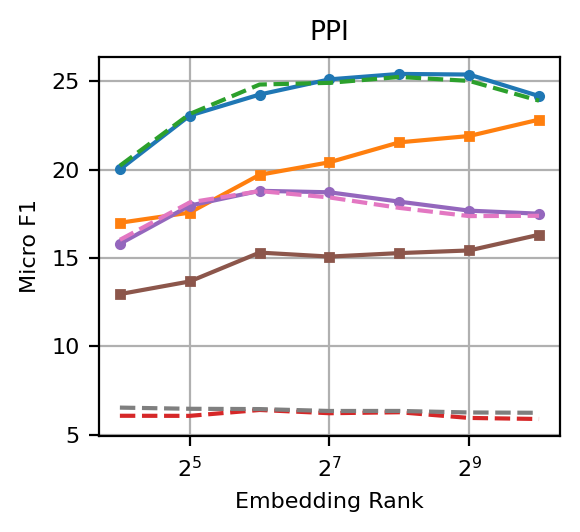}\hfill
\includegraphics[width=0.32\linewidth]{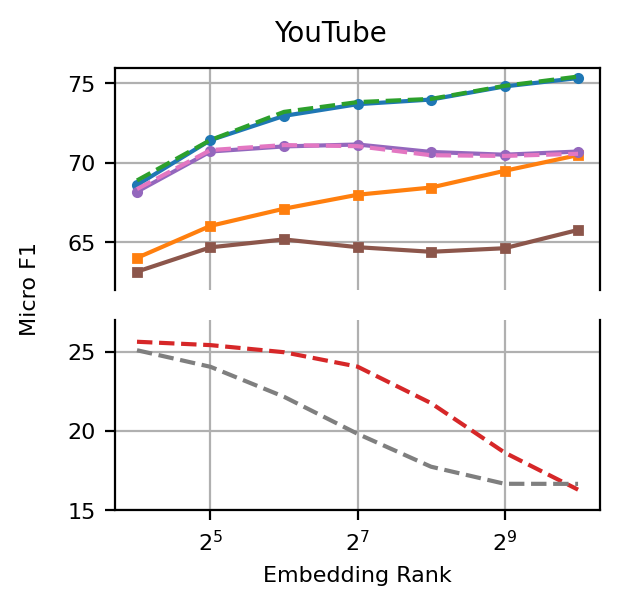} 
\caption{Multi-label classification using embeddings from reconstructed networks. Performance when using embeddings from a random graph is included as a baseline.}
\label{fig:real_class}
\end{figure}

In Figure~\ref{fig:conductancesmall}, we plot the relative errors for the conductances of the five most populous communities of the networks under consideration. We see that the conductance of ground-truth communities is generally preserved in the reconstructed networks,
with the error becoming negligible after rank $2^7=128$, an embedding rank which is often used in practice. This finding is intuitive -- since NetMF embeddings are used for node classification and community detection, it is to be expected that they preserve community structure.
  
\spara{Node classification.} In a typical classification setting for a graph $G$, when we know only a fraction of the labels of its nodes and want to infer the rest, we can use a low-dimensional embedding of its nodes as our feature matrix and employ a linear classifier to infer the labels for the remaining nodes. While our reconstructed networks $\tilde G$ differ from $G$ edge-wise, they have similar low-dimensional NetMF embeddings. As another indicator of the preservation of community structure, we measure the performance in this node classification task when using the embeddings $\mathcal{E}(\tilde G)$ as our feature matrix in place of $\mathcal{E}(G)$. 
We report the performance of two embeddings made from reconstructed networks: by applying NetMF to $\tilde G$ before and after binarizing its edges as described in Section \ref{subsec:gradalg}.
 
Our classification setting is the same as that of \cite{QiuDongMa:2018}: we use a one-vs-rest logistic regression classifier, sampling
a certain portion of the nodes as the training set. We repeat this sampling procedure $10$ times and report the mean micro F1 scores. We also repeat the experiments as we vary the embedding dimensionality $k$ and as we change the ratio of labeled examples from 10\% to 90\%.

As shown in Figure~\ref{fig:real_class}, when we use $\mathcal{E}(\tilde G)$ generated from the non-binarized (i.e., expected) $\tilde G$ as the input to our
logistic regression classifier, we achieve almost equal performance to when we use the true embedding $\mathcal{E}(G)$. This finding can be interpreted in two ways. First, it shows that the low error observed in Figure~\ref{fig:frobenius_embedding} (left) extends beyond the Frobenius norm metric, to the perhaps more directly meaningful metric of comparable performance in classification. Second, it makes clear that losing local connectivity properties in the inversion process (like total triangle count and the existence of specific edges) does not significantly effect classification performance. 
The reconstructed networks seem to preserve more global properties that are important for node classification, like community structure.  

While binarization does not significantly affect other metrics used to compare $\tilde G$ to $G$ (e.g., adjacency error, triangles), the classification task seems to be more
sensitive, as performance falls when we use the embedding for the binarized $\tilde G$. It is an interesting open direction to investigate this phenomenon, and generally how the low-dimensional embeddings of a probabilistic adjacency matrix change when that matrix is sampled to produce an unweighted graph. 

\spara{Synthetic graphs.} We repeat the above experiments using several synthetic networks produced by the stochastic block model (SBM)~\citep{abbe2015exact}. This random graph model assigns each node to a single cluster, and an edge between two nodes appears with probability $p_{in}$ if the nodes belong to the same cluster and $p_{out}$ otherwise, where generally it sets $p_{out} < p_{in}$. 
The configurations are summarized in Table~\ref{tab:sbm_config}. All networks have 1000 nodes, and, within each network, each cluster has the same size.
\begin{table}[h]
\centering
\begin{tabular}{llll}
        \toprule
        \textbf{Name} & \textbf{\# of Clusters} & \textbf{$p_{in}$} & \textbf{$p_{out}$} \\
        \midrule
        SBM 1 & 4 & 0.1 & 0.02  \\
        SBM 2 & 2 & 0.06 & 0.015 \\
        SBM 3 & 2 & 0.1 & 0.055 \\
        SBM 4 & 2 & 0.1 & 0.01 \\
        SBM 5 & 2 & 0.07 & 0.04 \\
        \bottomrule
\end{tabular}
\caption{\label{tab:sbm_config} Configuration of SBM networks; all networks have 1000 nodes.}
\end{table}

\begin{figure}
\centering
\includegraphics[width=0.40\linewidth]{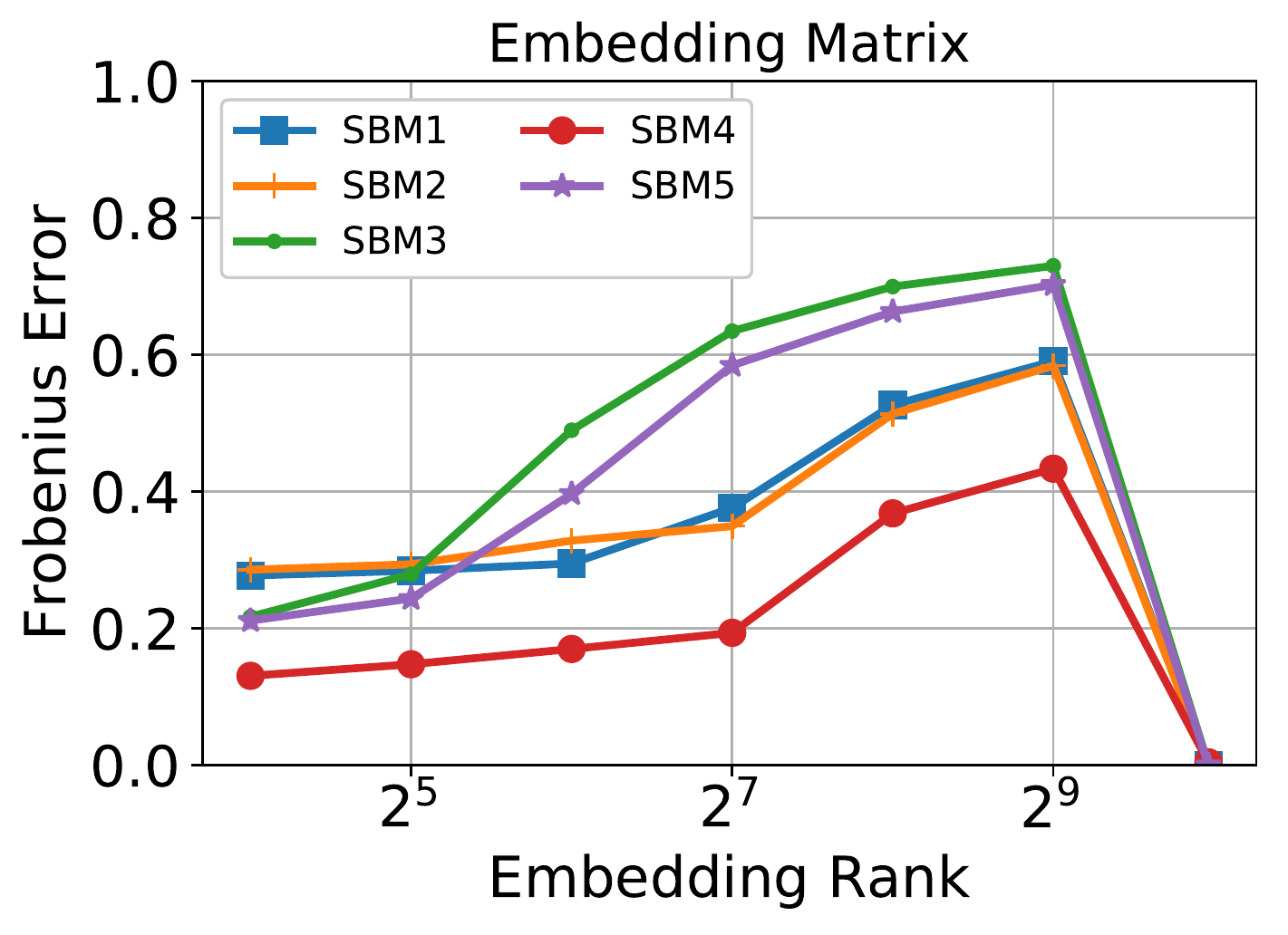} \quad
\includegraphics[width=0.40\linewidth]{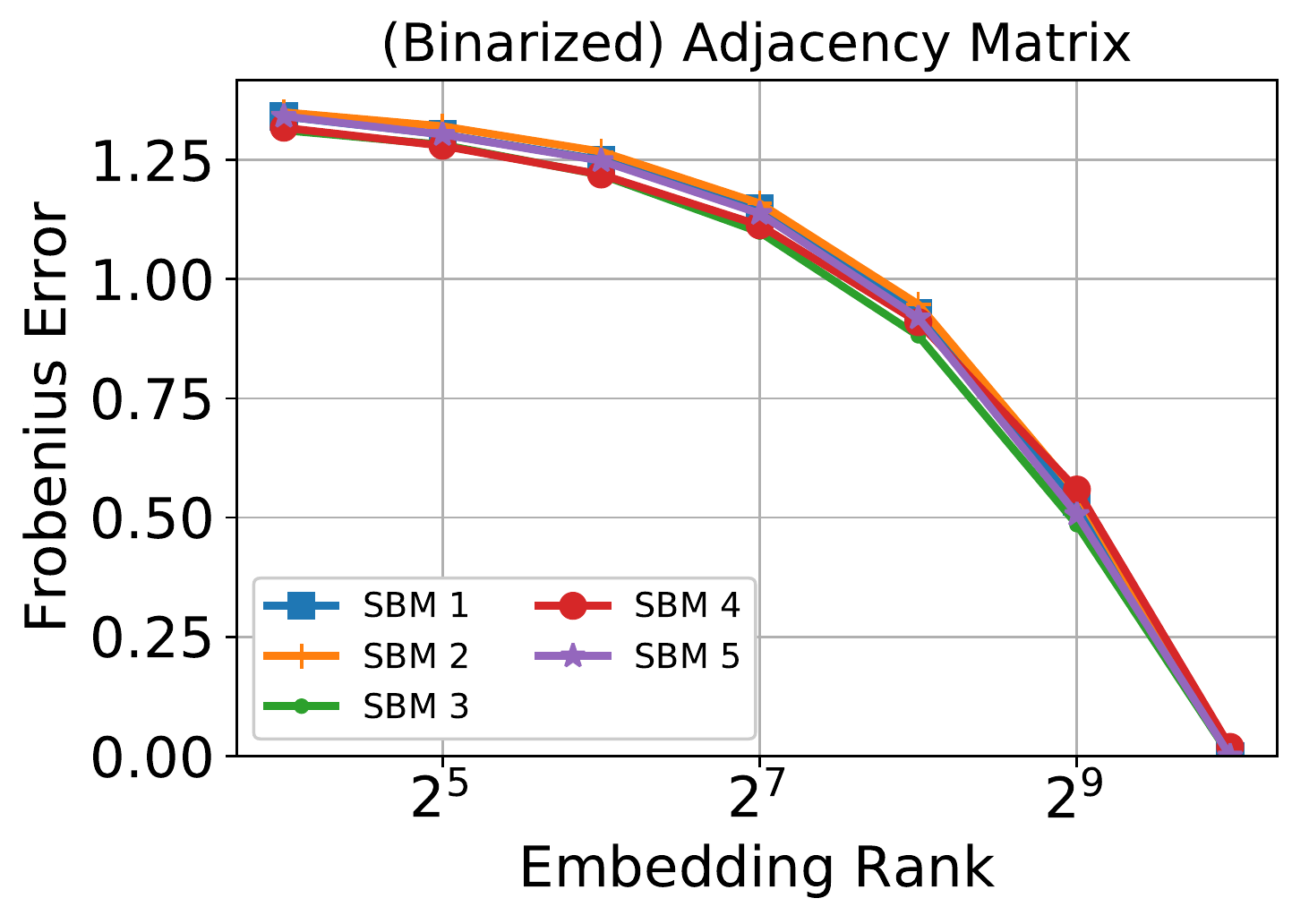} 
\caption{Relative Frobenius error for the low-rank PPMI matrices of reconstructions of the synthetic SBM networks (left) and the binarized adjacency matrix (right).}
\label{fig:frobenius_embedding_sbm}
\end{figure}

\begin{figure}
    \centering
    \includegraphics[width=0.40\linewidth]{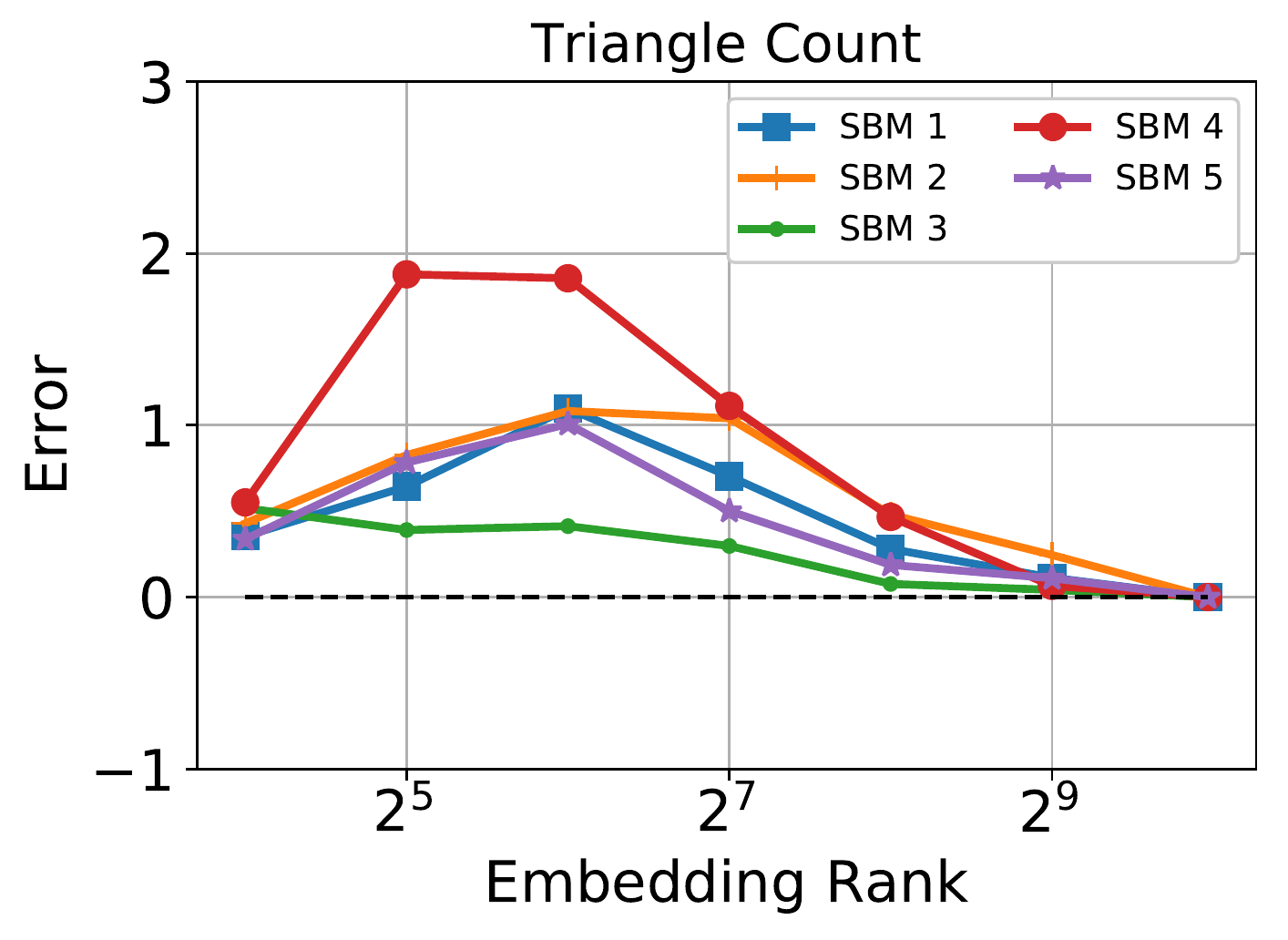} \quad
    \includegraphics[width=0.40\linewidth]{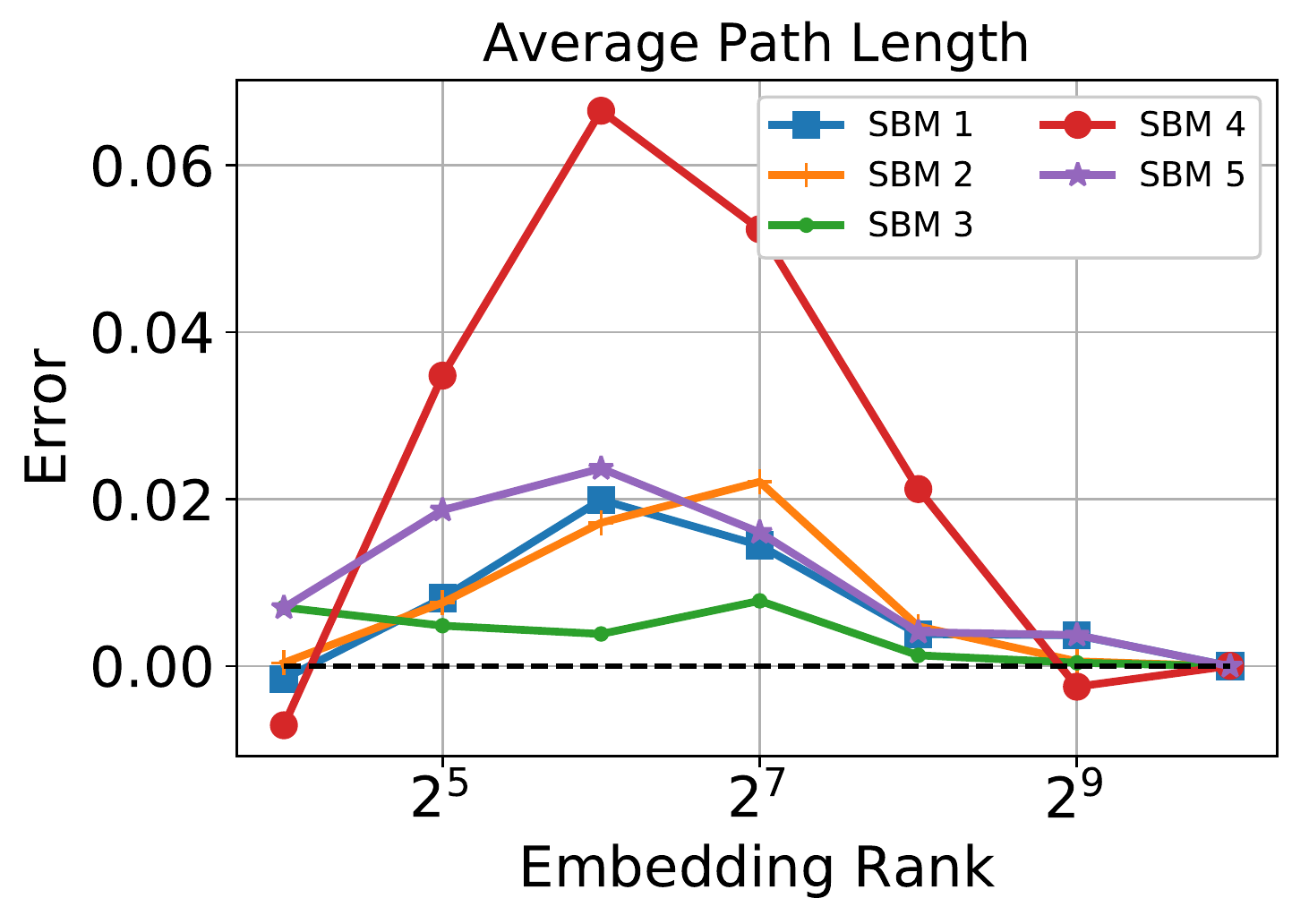} 
    \caption{Graph reconstruction errors for synthetic SBM networks. Relative error for the number of triangles (left) and for the average path length (right).}
    \label{fig:apl_triangles_sbm}
\end{figure}

\begin{figure}[h]
\centering
\includegraphics[width=0.32\linewidth]{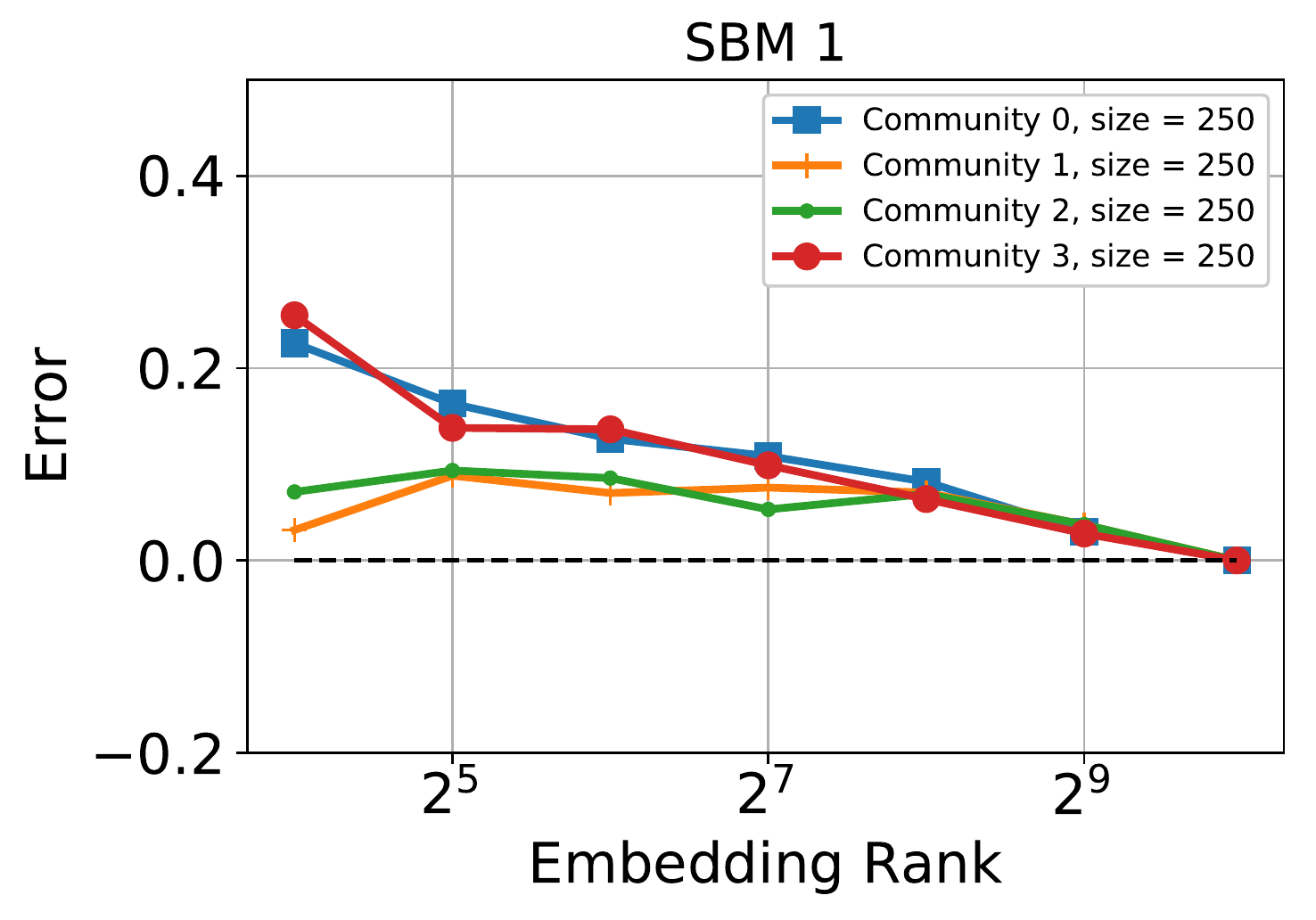} \hfill
\includegraphics[width=0.32\linewidth]{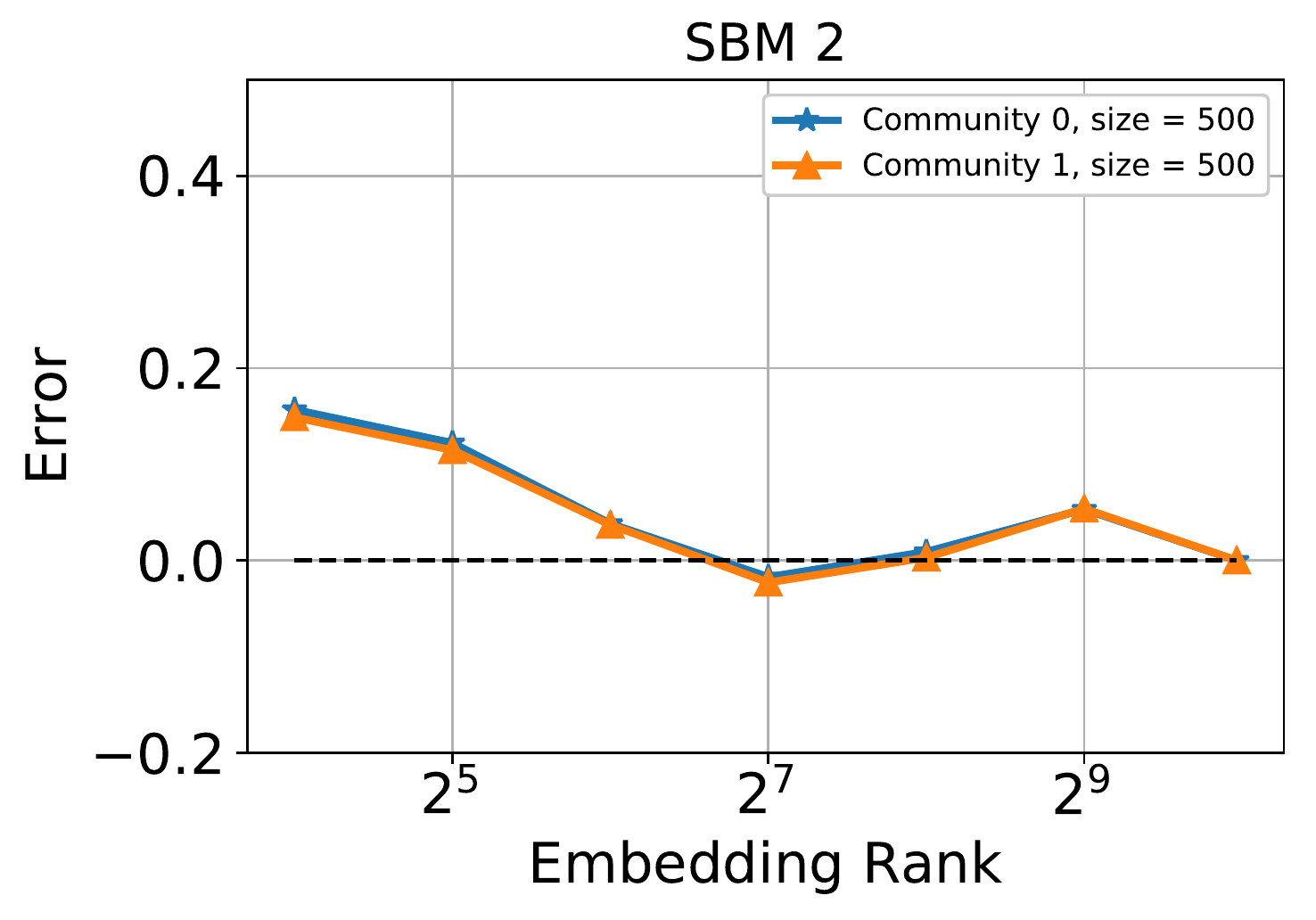} \hfill
\includegraphics[width=0.32\linewidth]{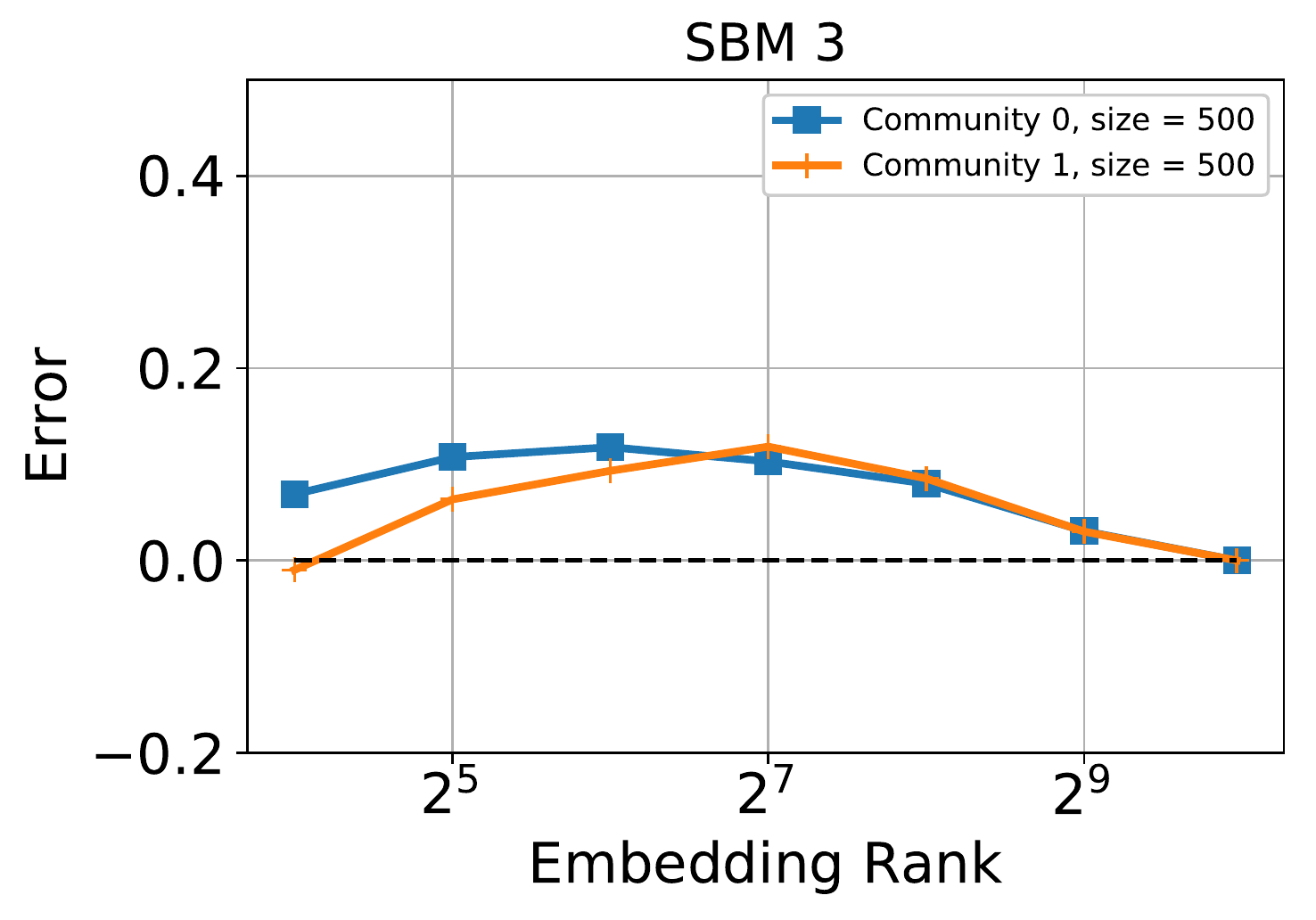}\\
\includegraphics[width=0.32\linewidth]{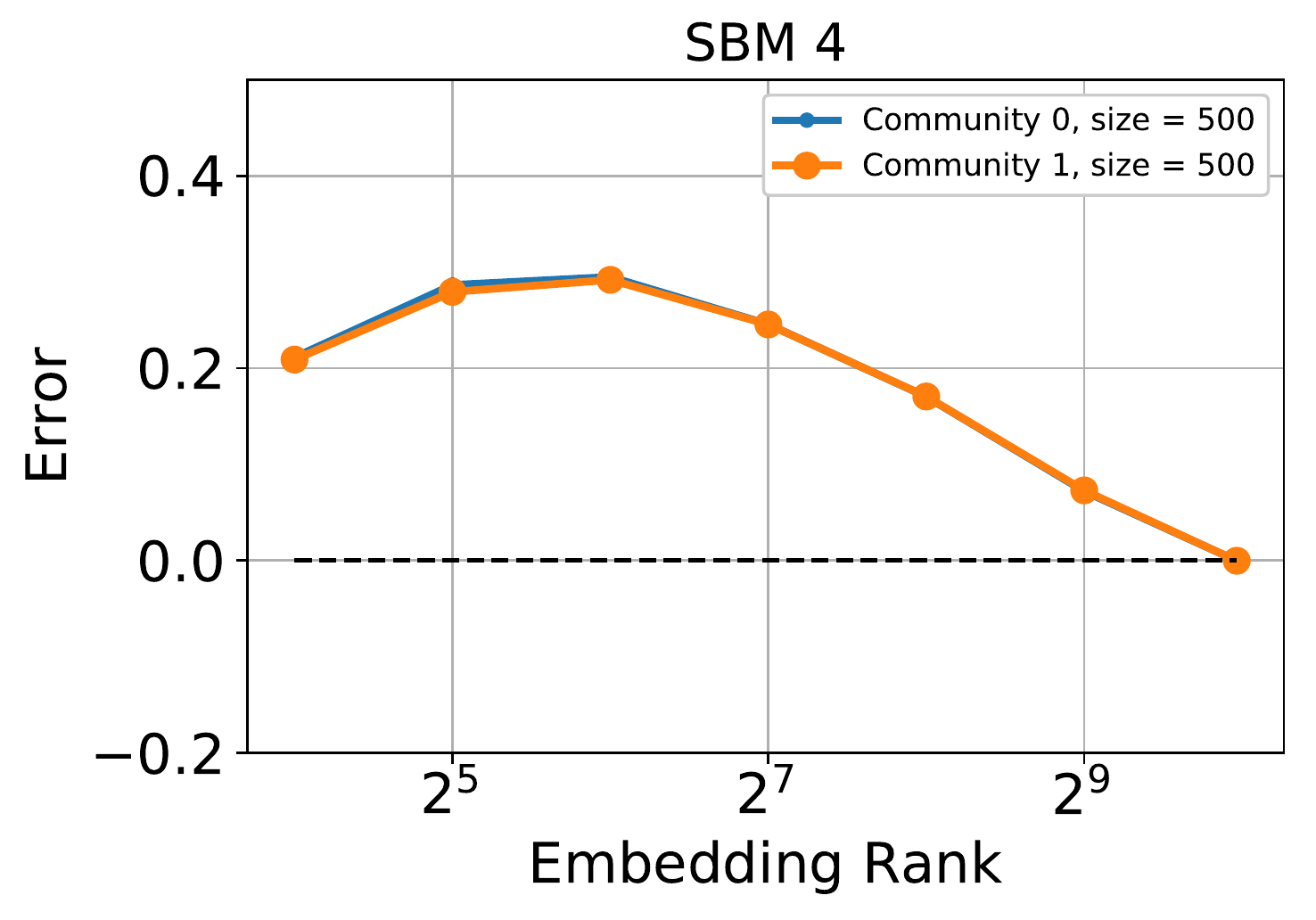} \quad
\includegraphics[width=0.32\linewidth]{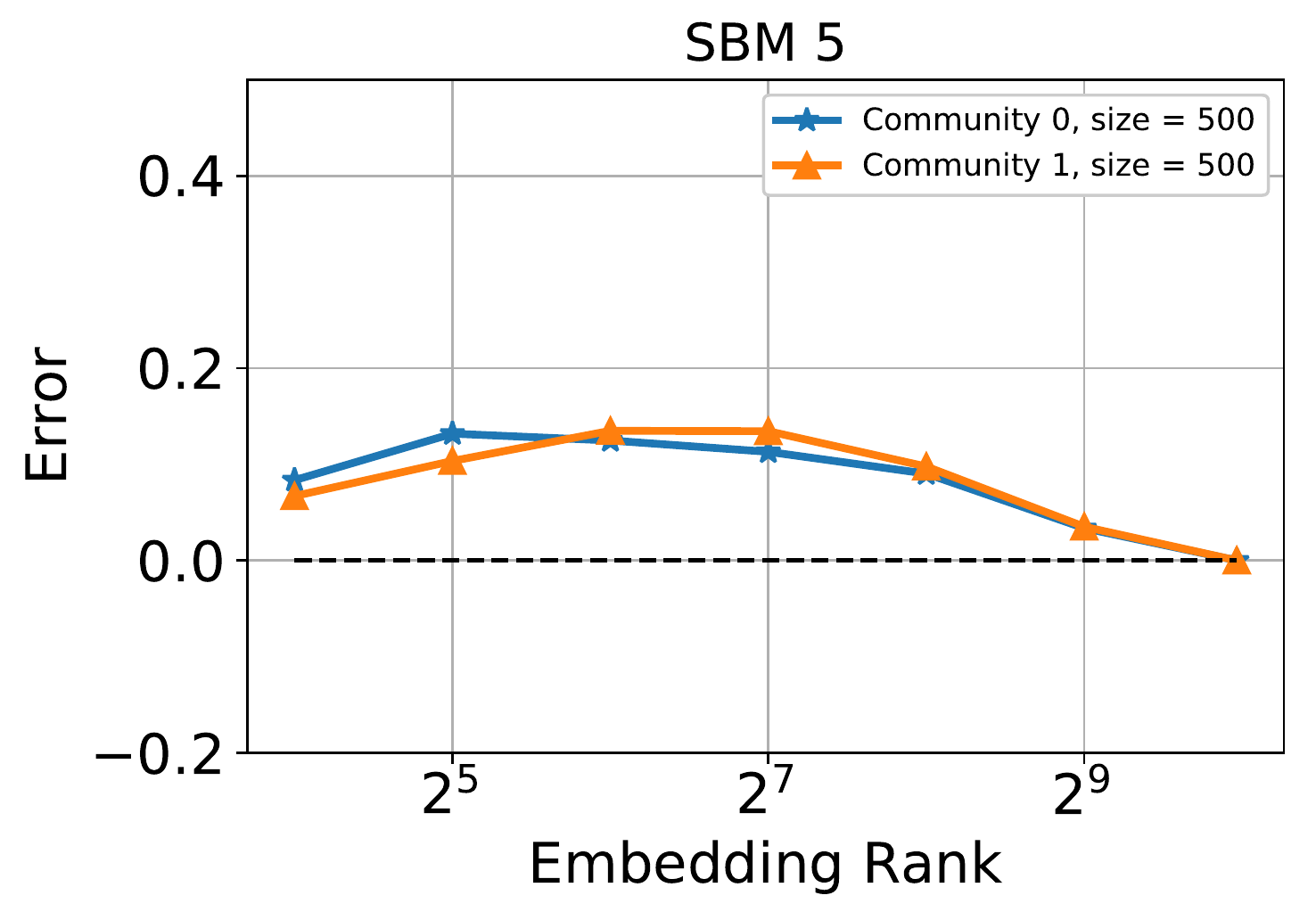}
\caption{Relative error for the conductances of the five most populous communities for each synthetic SBM network.}
\label{fig:conductance_sbm}
\end{figure}

\begin{figure}[h]
\centering
\includegraphics[width=0.32\linewidth]{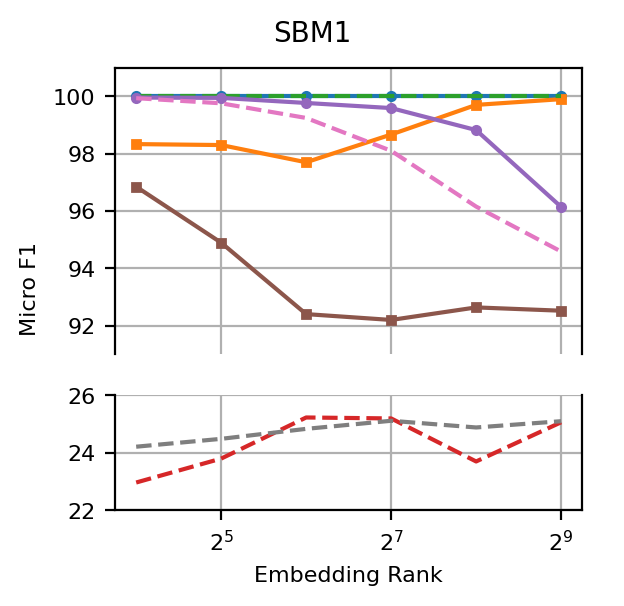} \quad
\includegraphics[width=0.32\linewidth]{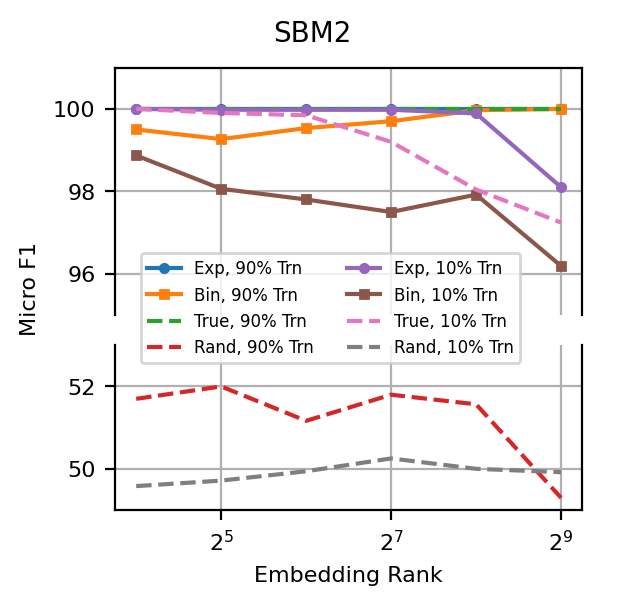} 
\caption{Multi-label classification using embeddings from reconstructions of two of the synthetic SBM networks.}
\label{fig:sbm_class}
\end{figure}

As with the real-world networks, we include plots for the error of the NetMF embedding matrix and the binarized adjacency matrix; (Figure~\ref{fig:frobenius_embedding_sbm}; the error of triangles count, and average path length (Figure~\ref{fig:apl_triangles_sbm}); the error of the conductances of the top communities (Figure~\ref{fig:conductance_sbm}); and the node classification performance using embeddings made from the reconstructed networks (Figure~\ref{fig:sbm_class}). For the node classification task, each node is a member of a single ground-truth community which corresponds to its cluster in the SBM. 

The results here largely match those of the real-world networks: the networks recovered by applying NetMF embedding inversion differ substantially from the true networks in terms of adjacency matrix and triangle count. However, we observe that community structure is well preserved -- see Figure~\ref{fig:intro_matshow} for a visual depiction.

Finally, we note that when our input is the full rank PPMI matrix (i.e., $k = n$), we succeed in reconstructing $G$ exactly (i.e., $\tilde G = G$) for the SBM networks. This further supports the message of Theorem~\ref{thm:invert} that, when embedding dimensionality is sufficiently high, node embeddings can be exactly inverted. However, at low dimensions, the embeddings seem to capture some important global properties, including community structure, while washing out more local structure. 
 
\section{Conclusion}
\label{sec:concl}
Node embeddings have been instrumental in achieving state-of-the-art results for graph-based machine learning tasks. Our work is a step towards a deeper understanding of why this is the case. We initiate the study of node embedding inversion as a tool to probe the information encoded in these embeddings. For the NetMF embedding method, we propose two approaches based on different techniques, and we show that the inversion problem can be effectively solved. Building on this, we show that while these embeddings seem to wash out local information in the underlying graph, they can be inverted to recover a graph with similar community structure to the original. Two interesting questions are whether our framework can be extended beyond the NetMF method, and whether we can formalize our empirical findings mathematically. We believe that our framework can be extended to the broader family of node embeddings that are based on low-rank factorization of graph similarity matrices. We hope that comparing the invertibility  of such embeddings can shed light on the differences and similarities between them. 


\bibliographystyle{plainnat}
\bibliography{main}

\clearpage
\appendix

\label{sec:appendix} 
\section{Appendix}\label{sec:appendix}

\subsection{Recovery of Degrees from Limiting PMI} \label{app:degrees}
For an undirected graph $G$ with adjacency matrix $A$ and unnormalized Laplacian $L$, let $\bm{d}$ be the vector with $i^{th}$ entry equal to the $i^{\text{th}}$ node's degree and $\bm{d}^{1/2}$ be its entrywise square root. Note that
\[ \bar L \bm{d}^{1/2} = D^{-1/2} L D^{-1/2} \bm{d}^{1/2} = D^{-1/2} L \bm{1} = \bm{0} \]
 since the all-ones vector $\bm{1}$ is in the null space of the unnormalized Laplacian $L$.

Suppose we have the limiting PMI matrix $M_\infty$ and the graph volume $v_G$. We subtract the all-ones matrix $J$ from $M_\infty$ and multiply by $\bm{d} / v_G$:
\begin{align*}
(M_\infty - J) (\bm{d} / v_G)
&=
v_G \cdot D^{-1/2} (\bar L^+ - I) D^{-1/2} (\bm{d}/v_G) \\
&= D^{-1/2} \bar L^+ d^{1/2} -  D^{-1/2} I \bm{d}^{1/2}\\
& = 0 - \bm{1} = -\bm{1}.
\end{align*}

Thus, if we solve the linear system $(M_\infty - J) \bm{x} = -\bm{1}$ for $\bm{x}$, we should get $\bm{x} = \bm{d}/v_G$, from which we can determine all nodes' degrees. Note that without $v_G$, we can still recover the degrees up to a constant factor.

The only issue with the above approach occurs when $(M_\infty - J)$ is singular and the linear system does not have a unique solution. $(M_\infty - J)$ is singular iff $(\bar L^+ - I)$ is singular, and this only occurs when $\bar L^+$ and hence $\bar L$ has an eigenvalue equal to $1$. $\bar L = I - D^{-1/2} A D^{-1/2}$, so this requires that $D^{-1/2} A D^{-1/2}$ has a zero eigenvalue. Thus, $\bar L^+ - I$ is singular exactly when $A$ is singular.

\end{document}